\documentclass[10pt,twocolumn,letterpaper]{article}

\usepackage{iccv}
\usepackage{times}
\usepackage{epsfig}
\usepackage{graphicx}
\usepackage{amsmath}
\usepackage{amssymb}
\usepackage{subcaption}
\usepackage{float}
\usepackage{MnSymbol,bbding,pifont}
\usepackage{multirow}
\usepackage{siunitx}
\usepackage{hhline}
\usepackage[normalem]{ulem} 
\usepackage{xcolor}
\definecolor{deemph}{gray}{0.6}
\newcommand{\gc}[1]{\textcolor{deemph}{#1}}
\newcommand{\implus}[1]{\textcolor{black}{#1}}
\newcommand{\doubleChecked}[1]{\textcolor{black}{#1}}

\usepackage{booktabs}
\usepackage{makecell}
\usepackage{tabularx} 
\usepackage{colortbl}
\usepackage{comment}
\usepackage{CJKutf8}
\usepackage{graphicx}
\graphicspath{{i/}}
\usepackage{caption, subcaption}
\DeclareCaptionFont{ninept}{\fontsize{9pt}{11pt}\selectfont #1}
\newcommand{\executeiffilenewer}[3]{%
 \ifnum\pdfstrcmp{\pdffilemoddate{#1}}%
 {\pdffilemoddate{#2}}>0%
 {\immediate\write18{#3}}\fi%
}

\usepackage[pagebackref=true,breaklinks=true,letterpaper=true,colorlinks,bookmarks=false]{hyperref}

\iccvfinalcopy 

\ificcvfinal\fi

\begin{document}

\title{Large Selective Kernel Network for Remote Sensing Object Detection}

\author{Yuxuan Li,\ \ Qibin Hou,\ \ Zhaohui Zheng,\ \ Ming-Ming Cheng,\ \ Jian Yang\ \ and\ \ Xiang Li\thanks{Corresponding author. Team site: \href{https://github.com/IMPlus-PCALab}{https://github.com/IMPlus-PCALab}
} \\
IMPlus@PCALab \& TMCC, CS, Nankai University\\
{\tt\small yuxuan.li.17@ucl.ac.uk, andrewhoux@gmail.com,} \\ 
{\tt\small   \{zh\_zheng, cmm, csjyang, xiang.li.implus\}@nankai.edu.cn}
}


\ificcvfinal\thispagestyle{empty}\fi
\maketitle

\begin{abstract}
  Recent research on remote sensing object detection has largely focused on improving the representation of oriented bounding boxes but has overlooked the unique prior knowledge presented in remote sensing \implus{scenarios.} Such prior knowledge can be useful because tiny remote sensing objects may be mistakenly detected without referencing a sufficiently long-range context, and the long-range context required by different types of objects can vary. In this paper, we take these priors into account and propose the Large Selective Kernel Network (LSKNet). LSKNet can dynamically adjust its large spatial receptive field to better model the ranging context of various objects in remote sensing scenarios. To the best of our knowledge, this is the first time that large and selective kernel mechanisms have been explored in the field of remote sensing object detection. Without bells and whistles, LSKNet sets new state-of-the-art scores on standard benchmarks, i.e., HRSC2016 (98.46\% mAP), DOTA-v1.0 (81.85\% mAP) and FAIR1M-v1.0 (47.87\% mAP). Based on a similar technique, we rank 2nd place in 2022 the Greater Bay Area International Algorithm Competition. Code is available at \href{https://github.com/zcablii/Large-Selective-Kernel-Network}{https://github.com/zcablii/Large-Selective-Kernel-Network}.

\end{abstract}

\section{Introduction}

\doubleChecked{Remote sensing object detection~\cite{zaidi_survey_2022} is a field of computer vision that focuses on identifying and locating objects of interest in aerial images, such as vehicles or aircraft. 
In recent years, one mainstream trend is to generate bounding boxes that accurately fit the orientation of the objects being detected, rather than simply drawing horizontal boxes around them.
Consequently, a significant amount of research has focused on improving the representation of oriented bounding boxes for remote sensing object detection. This has largely been achieved through the development of specialized detection frameworks, such as RoI Transformer~\cite{ding_learning_2019}, Oriented R-CNN~\cite{xie_oriented_2021} and R3Det~\cite{yang_r3det_nodate}, as well as techniques for oriented box encoding, such as gliding vertex~\cite{xu_gliding_2021} and midpoint offset box encoding~\cite{xie_oriented_2021}. Additionally, a number of loss functions, including GWD~\cite{yang_rethinking_2021}, KLD~\cite{yang_learning_2021} and Modulated Loss~\cite{qian_learning_2021}, have been proposed to further enhance the performance of these approaches.
}

\begin{figure}[t]
\begin{center}
\includegraphics[width=1\linewidth]{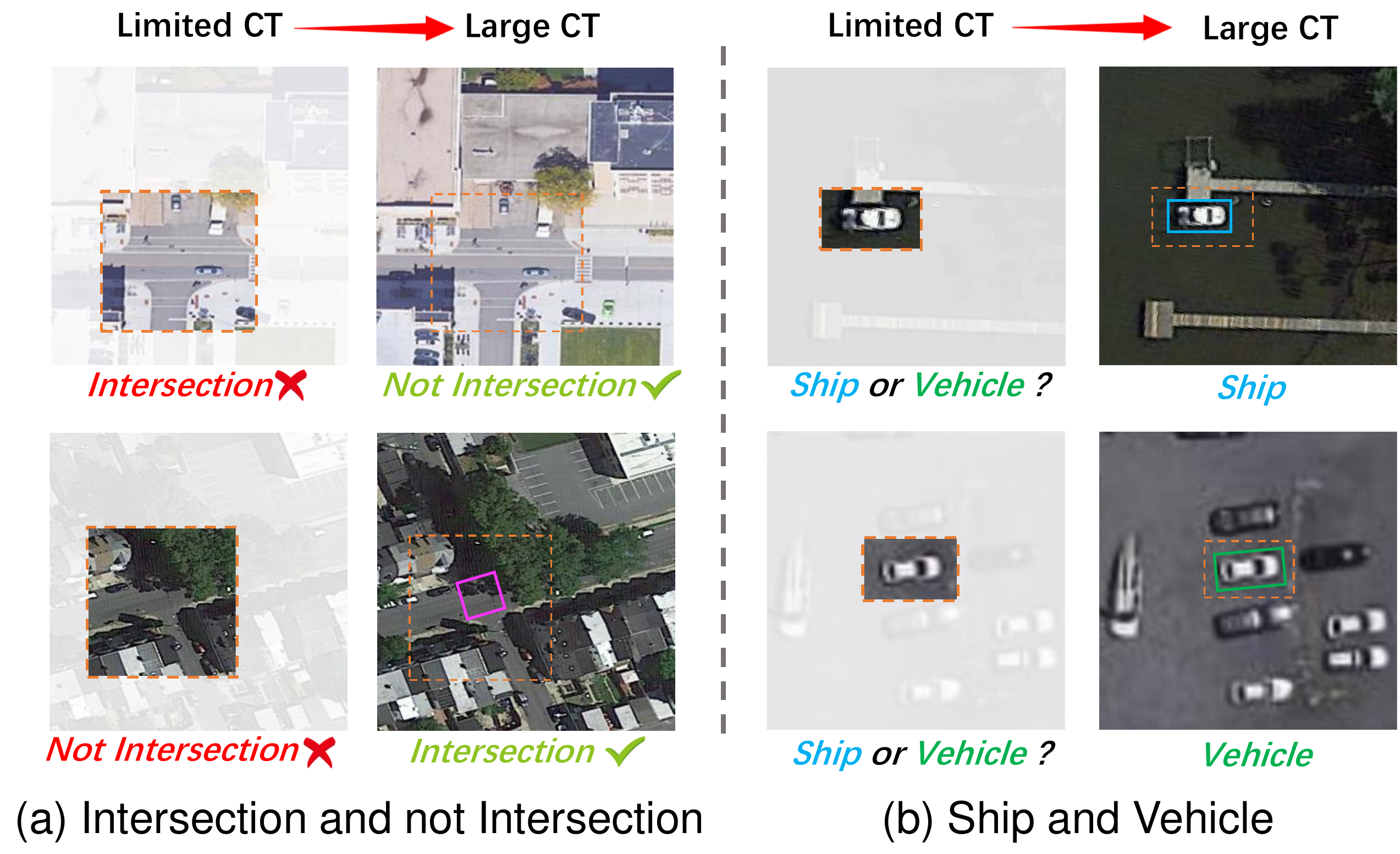}
\end{center}
    \vspace{-16pt}
   \caption{Successfully detecting remote sensing objects requires the use of a wide range of contextual information. Detectors with a limited receptive field may easily lead to incorrect detection results. ``CT'' stands for Context.}
\label{fig:finding_1}
\vspace{-12pt}
\end{figure}
\doubleChecked{
However, despite these advances, relatively few works have taken into account the strong prior knowledge that exists in remote sensing images. Aerial images are typically captured from a bird's eye view at high resolutions.
In particular, most objects in aerial images may be small in size and difficult to identify based on their appearance alone. Instead, the successful recognition of these objects often relies on their context, as the surrounding environment can provide valuable clues about their shape, orientation, and other characteristics. According to an analysis of mainstream remote sensing datasets, we identify two important priors:
}

        \textbf{(1) Accurate detection of objects in remote sensing images often requires a wide range of contextual information.} 
        \doubleChecked{As illustrated in Fig.~\ref{fig:finding_1}(a), the limited context used by object detectors in remote sensing images can often lead to incorrect classifications. In the upper image, for example, the detector may classify the junction as an intersection due to its typical characteristics, but in reality, it is not an intersection. Similarly, in the lower image, the detector may classify the junction as not being an intersection due to the presence of large trees, but again, this is incorrect. These errors can occur because the detector is only considering a limited amount of contextual information in the immediate vicinity of the objects. A similar scenario can be also observed in the example of ships and vehicles in Fig.~\ref{fig:finding_1}(b).}
        
\begin{figure}[t]
\begin{center}
\includegraphics[width=0.85\linewidth]{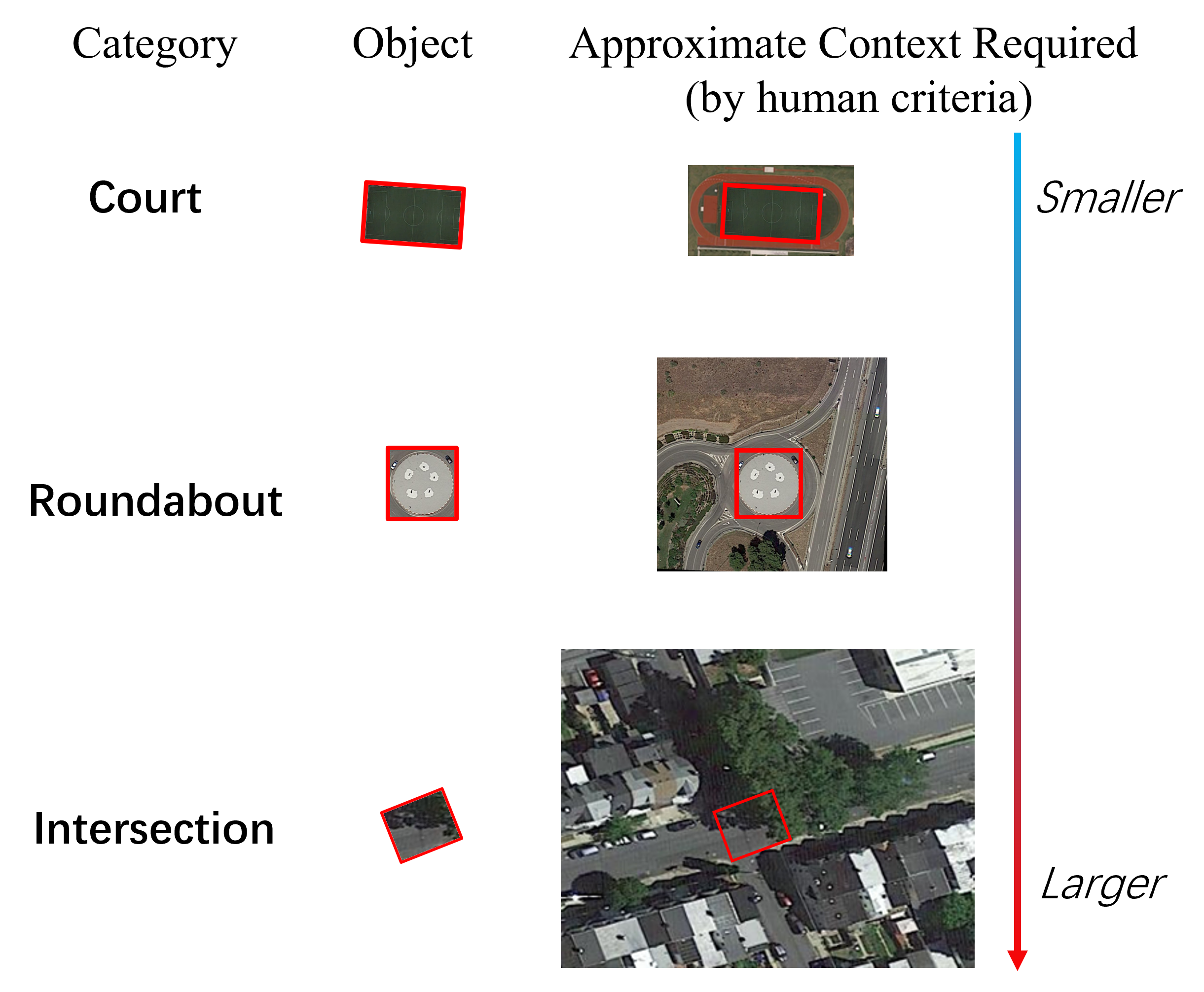}
\end{center}
    \vspace{-16pt}
   \caption{The wide range of contextual information required for different object types is very different by human criteria. The objects with red boxes are the exact ground-truth annotations.}
\label{fig:finding_2}
\vspace{-12pt}
\end{figure}

        \textbf{(2) The wide range of contextual information required for different object types is very different. }
        \doubleChecked{As shown in Fig.~\ref{fig:finding_2}, the amount of contextual information required for accurate object detection in remote sensing images can vary significantly depending on the type of object being detected.
        For example, Soccer-ball-field may require relatively less extra contextual information because of the unique distinguishable court borderlines.
        In contrast, Roundabouts may require a larger range of context information in order to distinguish between gardens and ring-like buildings. Intersections, especially those that are partially covered by trees, often require an extremely large receptive field due to the long-range dependencies between the intersecting roads. This is because the presence of trees and other obstructions can make it difficult to identify the roads and the intersection itself based on appearance alone.
        Other object categories, such as bridges, vehicles, and ships, may also require different scales of the receptive field in order to be accurately detected and classified.}

 \doubleChecked{To address the challenge of accurately detecting objects in remote sensing images, which often require a wide and dynamic range of contextual information, we propose a novel approach called Large Selective Kernel Network (LSKNet). 
 Our approach involves dynamically adjusting the receptive field of the feature extraction backbone in order to more effectively process the varying wide context of the objects being detected. This is achieved through a \implus{spatial} selective mechanism, which weights the features processed by a sequence of large depth-wise kernels efficiently and then spatially merges them. The weights of these kernels are determined dynamically based on the input, allowing the model to adaptively use different large kernels and adjust the receptive field for each target in space as needed.}

\doubleChecked{To the best of our knowledge, our proposed LSKNet is the first to investigate and discuss the use of large and selective kernels for remote sensing object detection. Despite its simplicity, our model achieves state-of-the-art performance on three popular datasets: HRSC2016 (98.46\% mAP), DOTA-v1.0 (81.64\% mAP), and FAIR1M-v1.0 (47.87\% mAP), surpassing previously published results. Furthermore, we demonstrate that our model's behaviour exactly aligns with the aforementioned two priors, which in turn verifies the effectiveness of the proposed mechanism. }

\begin{figure*}[!htb]
\hspace{0.1in}
  \begin{subfigure}[b]{0.2\textwidth}
    \includegraphics[page=1, width=\linewidth]{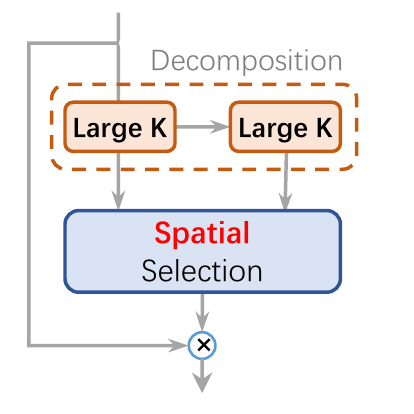}
    \caption{LSK (\textbf{ours})}
    \label{fig:convnext}
  \end{subfigure}
  \hspace{0.15in}
\unskip\ \color{gray} \vrule depth -5ex height 26ex\
\hspace{0.15in}
  \begin{subfigure}[b]{0.2\textwidth}
    \includegraphics[page=2, width=\linewidth]{imgs/arch_comp_cropped.pdf}
    \caption{ResNeSt}
    \label{fig:sk}
  \end{subfigure}
  \hfill 
  \begin{subfigure}[b]{0.2\textwidth}
    \includegraphics[page=3, width=\linewidth]{imgs/arch_comp_cropped.pdf}
    \caption{SCNet}
    \label{fig:resnest}
  \end{subfigure}
  \hfill 
  \begin{subfigure}[b]{0.2\textwidth}
    \includegraphics[page=4, width=\linewidth]{imgs/arch_comp_cropped.pdf}
    \caption{SKNet}
    \label{fig:SCNet}
  \end{subfigure}
  \hfill 
\vspace{-8pt}
\caption{Architectural comparison between our proposed LSK module and other representative selective mechanism modules. K: Kernel.}
\label{fig:arch_compare}
\vspace{-8pt}
\end{figure*}

\section{Related Work}

\subsection{Remote Sensing Object Detection Framework}

High-performance remote sensing object detectors often rely on the RCNN~\cite{frcnn} framework, which consists of a region proposal network and regional CNN detection heads. 
Several variations on the RCNN framework have been proposed in recent years. The two-stage RoI transformer~\cite{ding_learning_2019} uses fully-connected layers to rotate candidate horizontal anchor boxes in the first stage, and then features within the boxes are extracted for further regression and classification. SCRDet~\cite{yang_scrdet_2019} uses an attention mechanism to reduce background noise and improve the modelling of crowded and small objects. Oriented RCNN~\cite{xie_oriented_2021} and Gliding Vertex~\cite{xu_gliding_2021} introduce new box encoding systems to address the instability of training losses caused by rotation angle periodicity. 
Some approaches~\cite{law2018cornernet, zhou2019objects, wang_learning_2021} treat remote sensing detection as a point detection task~\cite{yang2017stacked}, providing an alternative way of addressing remote sensing detection problems.

Rather than relying on the proposed anchors, one-stage detection frameworks classify and regress oriented bounding boxes directly from grid densely sampled anchors. The one-stage S$^2$A network~\cite{han_align_2020} extracts robust object features via oriented feature alignment and orientation-invariant feature extraction. DRN~\cite{pan_dynamic_2020}, on the other hand, leverages attention mechanisms to dynamically refined the backbone's extracted features for more accurate predictions. In contrast with Oriented RCNN and Gliding Vertex, RSDet~\cite{qian_learning_2021} addresses the discontinuity of regression loss 
by introducing a modulated loss.  AOPG~\cite{cheng_anchor-free_2022} and R3Det~\cite{yang_r3det_nodate} adopt a progressive regression approach, refining bounding boxes from coarse to fine granularity. In addition to CNN-based frameworks, AO2-DETR~\cite{dai_ao2-detr_2022} introduces a transformer-based detection framework, DETR~\cite{carion2020end}, into remote sensing detection tasks, which brings more research diversity.

While these approaches have achieved promising results in addressing the issue of rotation variance, they do not take into account the strong and valuable prior information presented in aerial images. Instead, our approach focuses on leveraging the large kernel and spatial selective mechanism to better model these priors, without modifying the current detection framework.

\subsection{Large Kernel Networks}
Transformer-based~\cite{vaswani2017attention} models, such as the Vision Transformer (ViT)~\cite{dosovitskiy2020image, sar_vit, wang_advancing_2022, 9531646, rs13030516}, Swin transformer~\cite{liu2021swin, rs14061507, rs13234779, 9736956, rs13245100} and PVT~\cite{pvt} have gained popularity in computer vision due to their effectiveness in image recognition tasks. 
Research~\cite{Ranftl_2021_ICCV, DBLP, Zheng_2021_CVPR, luo_understanding_2016} has demonstrated that the large receptive field is a key factor in their success. In light of this, recent work has shown that well-designed convolutional networks with large receptive fields can also be highly competitive with transformer-based models. For example, ConvNeXt~\cite{liu2022convnet} uses 7$\times$7 depth-wise convolutions in their backbone, resulting in significant performance improvements in downstream tasks. In addition, RepLKNet~\cite{ding2022scaling} even uses a 31$\times$31 convolutional kernel via re-parameterization, achieving compelling performance. A subsequent work SLaK~\cite{liu2022more}, further expands the kernel size to 51$\times$51 through kernel decomposition and sparse group techniques. VAN~\cite{guo_visual_2022} introduces an efficient decomposition of large kernels as convolutional attention. Similarly, SegNeXt~\cite{guo_segnext_2022} and Conv2Former~\cite{hou_conv2former_2022} demonstrate that large kernel convolution plays an important role in modulating the convolutional features with a richer context. 

Despite the fact that large kernel convolutions have received attention in the domain of general object recognition, there has been a lack of research examining their significance in the specific field of remote sensing detection. As previously noted in the \textit{Introduction}, aerial images possess unique characteristics that make large kernels particularly well-suited for the task of remote sensing. As far as we are aware, our work represents the first attempt to introduce large kernel convolutions for the purpose of remote sensing and to examine their importance in this field.

\subsection{Attention/Selective Mechanism}

The attention mechanism is a simple and effective way to enhance neural representations for various tasks. The channel attention SE block~\cite{se_net} uses global average information to reweight feature channels, while spatial attention modules like GENet~\cite{genet}, GCNet~\cite{cao_gcnet_2019}, and SGE~\cite{li2022spatial} enhance a network's ability to model context information via spatial masks. CBAM~\cite{ferrari_cbam_2018} and BAM~\cite{Park2018BAMBA} combine both channel and spatial attention to make use of the advantages of both.

\begin{figure*}[t]
\begin{center}
\includegraphics[width=0.82\linewidth]{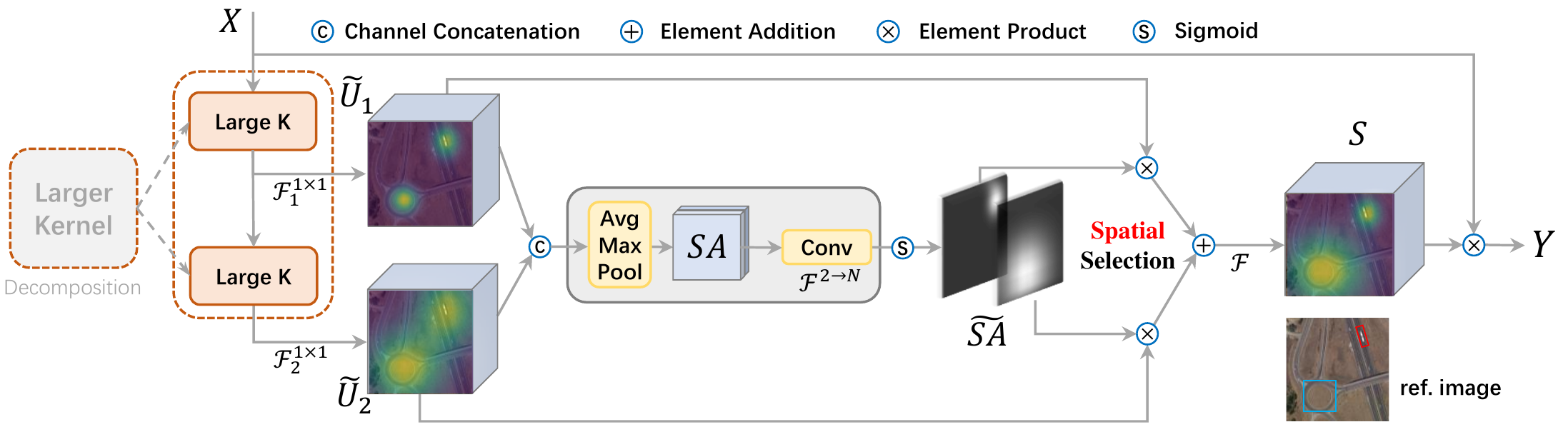}
\end{center}
    \vspace{-20pt}
   \caption{A conceptual illustration of LSK module.}
\label{fig:lsk_module}
\vspace{-8pt}
\end{figure*}

In addition to channel/spatial attention mechanisms, kernel selections are also a self-adaptive and effective technique for dynamic context modelling. CondConv~\cite{yang2019condconv} and Dynamic convolution~\cite{chen2020dynamic} use parallel kernels to adaptively aggregate features from multiple convolution kernels. SKNet~\cite{li2019selective} introduces multiple branches with different convolutional kernels and selectively combines them along the channel dimension. ResNeSt~\cite{resnest} extends the idea of SKNet by partitioning the input feature map into several groups. Similarly to the SKNet, SCNet~\cite{liu_improving_2020} uses branch attention to capture richer information and spatial attention to improve localization ability. {Deformable Convnets~\cite{Zhu_Deformablev2, Dai_Deformable} introduce a flexible kernel shape for convolution units. }

\doubleChecked{Our approach bears the most similarity to SKNet~\cite{li2019selective}, however, there are \textbf{two key distinctions} between the two methods. Firstly, our proposed selective mechanism relies explicitly on a sequence of large kernels via decomposition, a departure from most existing attention-based approaches. Secondly, our method adaptively aggregates information across large kernels in the spatial dimension, rather than the channel dimension as utilized by SKNet. This design is more intuitive and effective for remote sensing tasks, because channel-wise selection fails to model the spatial variance for different targets across the image space. The detailed structural comparisons are listed in Fig.~\ref{fig:arch_compare}.
}

\section{Methods}

\subsection{LSKNet Architecture}

The overall architecture is built upon the recent popular structures~\cite{liu2022convnet,PVT_v2,guo_visual_2022,hou_conv2former_2022,MetaFormer} (refer to the details in Supplementary Materials (SM)) with a repeated building block. 
The detailed configuration of different variants of LSKNet \implus{used in this paper} is listed in Tab.~\ref{tab:model_variants}. Each LSKNet block consists of two residual sub-blocks: the Large Kernel Selection (LK Selection) sub-block and the Feed-forward Network (FFN) sub-block.
The core LSK module (Fig.~\ref{fig:lsk_module}) is embedded in the LK Selection sub-block. It consists of a sequence of large kernel convolutions and a spatial kernel selection mechanism, which would be elaborated on later.

\begin{table}[t]
\centering
\renewcommand\arraystretch{1.2} 
\resizebox{0.99\columnwidth}{!}{
\begin{tabular}{lllll}
\textbf{~~~~Model} & \{{$C_1$, $C_2$, $C_3$, $C_4$}\} & \{{$D_1$, $D_2$, $D_3$, $D_4$}\} & ~~~\#P \\ 
\Xhline{3\arrayrulewidth}
\textcolor{black}{$\star$} LSKNet-T          & \{32, 64, 160, 256\}            & ~~~~~\{3, 3, 5, 2\}     &    ~~4.3M    \\ 
$\star$ LSKNet-S          & \{64, 128, 320, 512\}          & ~~~~~\{2, 2, 4, 2\}          &  14.4M  \\ 
\end{tabular}
}
\vspace{-4pt}
\caption{\textbf{Variants of LSKNet used in this paper}. $C_i$: feature channel number; $D_i$: number of LSKNet blocks of each stage $i$.  }
\label{tab:model_variants}
\vspace{-4pt}
\end{table}

\begin{table}[t]
\renewcommand\arraystretch{1.2} 
\centering
\resizebox{0.80\columnwidth}{!}{
\begin{tabular}{c|l|cc} 
\multicolumn{1}{l|}{RF}                          & ($k$, $d$) sequence                               & \multicolumn{1}{l}{~~~\#P} & \multicolumn{1}{l}{~~FLOPs}  \\ 
\Xhline{3\arrayrulewidth}
\multirow{2}{*}{{23}} & (23, 1) & ~40.4K  & ~~42.4G \\ 
\cline{2-4}
& (5, 1) $\longrightarrow$ (7, 3)  & ~11.3K  & ~~11.9G \\ 
\hline
\multirow{2}{*}{{29}} & (29, 1) & ~60.4K  & ~~63.3G  \\ 
\cline{2-4}
& (3, 1) $\longrightarrow$  (5, 2) $\longrightarrow$ (7, 3) & ~11.3K  & ~~13.6G  \\
\end{tabular}
}
\vspace{-8pt}
\caption{\textbf{Theoretical efficiency comparisons of two representative examples} by expanding single large depth-wise kernel into a sequence, given channels being 64. 
$k$: kernel size; $d$: dilation.
}
\label{tab:decomposite_efficiency}
\vspace{-4pt}
\end{table}

\subsection{Large Kernel Convolutions}
\doubleChecked{According to the \textit{prior (2)} as stated in \textit{Introduction}, it is suggested to model a series of multiple long-range contexts for adaptive selection. Therefore, we propose to construct a larger kernel convolution by \emph{explicitly decomposing} it into a sequence of depth-wise convolutions with a large growing kernel and increasing dilation. Specifically, the expansion of the kernel size~$k$, dilation rate~$d$ and the receptive field $RF$, of the $i$-th depth-wise convolution in the series are defined as follows:} 
\vspace{-5pt}
\begin{equation}
    k_{i-1} \leq  k_i;~d_1 = 1,~ d_{i-1} < d_i \leq RF_{i-1} \text{,}
    \label{eqn:dr}
\end{equation}
\vspace{-16pt}
\begin{equation}
    RF_1 = k_1,~ RF_i = d_i(k_i-1) + RF_{i-1}\text{.}
    \label{eqn:rf}
    \vspace{-2pt}
\end{equation}

\noindent The increasing of kernel size and dilation rate ensure that the receptive field expands quickly enough. We set an upper bound on the dilation rate to guarantee that the dilation convolution does not introduce gaps between feature maps. For instance, we can decompose a large kernel into 2 or 3 depth-wise convolutions as in Tab.~\ref{tab:decomposite_efficiency}, which have a theoretical receptive field of 23 and 29, respectively.

\doubleChecked{There are two advantages of the proposed designs. First, it explicitly yields multiple features with various large receptive fields, which makes it easier for later kernel selection. Second, sequential decomposition is more efficient than simply applying a single larger kernel. As shown in Tab.~\ref{tab:decomposite_efficiency}, under the same resulted theoretical receptive field, our decomposition greatly reduces the number of parameters compared to the standard large convolution kernels.}
To obtain features with rich contextual information from different ranges for input $\mathbf{X}$, a series of decomposed depth-wise convolutions with different receptive fields are applied: 
\vspace{-4pt}
\begin{equation}
    \mathbf{U}_0 = \mathbf{X} \text{,~~~~}  \mathbf{U}_{i+1} = \mathcal{F}^{dw}_i(\mathbf{U}_i)\text{,}
    \label{eqn:2}    
    \vspace{-5pt}
\end{equation}
where $\mathcal{F}^{dw}_i(\cdot)$ are depth-wise convolutions with kernel $k_i$ and dilation $d_i$. Assuming there are $N$ decomposed kernels, each of which is further processed by a 1$\times$1 convolution layer $\mathcal{F}^{1 \times 1}(\cdot)$:
    \vspace{-5pt}
\begin{equation}
    \widetilde{\mathbf{U}}_i =  \mathcal{F}^{1 \times 1}_i(\mathbf{U}_i), \text{ for } i \text{ in } [1,N]\text{,}
    \label{eqn:3}
    \vspace{-5pt}
\end{equation}
allowing channel mixing for each spatial feature vector.
Then, a selection mechanism is proposed
to dynamically select kernels for various objects based on the multi-scale features obtained, which would be introduced next.

\subsection{Spatial Kernel Selection}
To enhance the network's ability to focus on the most relevant spatial context regions for detecting targets, we use a spatial selection mechanism to spatially select the feature maps from large convolution kernels at different scales. 
Firstly, we concatenate the features obtained from different kernels with different ranges of receptive field:
    \vspace{-4pt}
\begin{equation}
    \widetilde{\mathbf{U}} = [\widetilde{\mathbf{U}}_1;...;\widetilde{\mathbf{U}}_i] \text{,}
    \label{eqn:4}
    \vspace{-4pt}
\end{equation}
and then efficiently extract the spatial relationship by applying channel-based average and maximum pooling (denoted as $\mathcal{P}_{avg}(\cdot)$ and $\mathcal{P}_{max}(\cdot)$) to $\widetilde{\mathbf{U}}$: %
    \vspace{-4pt}
\begin{equation}
    \mathbf{SA}_{avg} = \mathcal{P}_{avg}(\widetilde{\mathbf{U}})\text{, \ }
    \mathbf{SA}_{max} = \mathcal{P}_{max}(\widetilde{\mathbf{U}})\text{,}
    \label{eqn:5}
    \vspace{-4pt}
\end{equation}
where $\mathbf{SA}_{avg}$ and $\mathbf{SA}_{max}$ are the average and maximum pooled spatial feature descriptors. 
To allow information interaction among different spatial descriptors, we concatenate the spatially pooled features and use a convolution layer $\mathcal{F}^{2 \rightarrow N}(\cdot)$ to transform the pooled features (with 2 channels) into $N$ spatial attention maps:
\vspace{-4pt}
\begin{equation}
    \widehat{\mathbf{SA}} = \mathcal{F}^{2 \rightarrow N}([\mathbf{SA}_{avg};\mathbf{SA}_{max}])\text{.}
    \label{eqn:6.2}
    \vspace{-4pt}
\end{equation}

\noindent 
For each of the spatial attention maps, $\widehat{\mathbf{SA}}_i$, a sigmoid activation function is applied to obtain the individual spatial selection mask for each of the decomposed large kernels:
\vspace{-4pt}
\begin{equation}
    \widetilde{\mathbf{SA}}_i = \sigma(\widehat{\mathbf{SA}}_i)\text{,}
    \label{eqn:7}
    \vspace{-4pt}
\end{equation}
\noindent where $\sigma(\cdot)$ denotes the sigmoid function. The features from the sequence of decomposed large kernels are then weighted by their corresponding spatial selection masks and fused by a convolution layer $\mathcal{F}(\cdot)$ to obtain the attention feature $\mathbf{S}$:
\vspace{-4pt}
\begin{equation}
    \mathbf{S} = \mathcal{F}(\sum_{i=1}^N {(\widetilde{\mathbf{SA}}_i \cdot \widetilde{\mathbf{U}}_i)})\text{.}
    \label{eqn:8}
    \vspace{-6pt}
\end{equation}

\noindent The final output of the LSK module is the element-wise product between the input feature $\mathbf{X}$ and $\mathbf{S}$, similarly in~\cite{guo_visual_2022,guo_segnext_2022,hou_conv2former_2022}:
\vspace{-4pt}
\begin{equation}
    \mathbf{Y} = \mathbf{X} \cdot \mathbf{S}\text{.}
    \label{eqn:9}
    \vspace{-6pt}
\end{equation}

\noindent Fig.~\ref{fig:lsk_module} shows a detailed conceptual illustration
of an LSK module where we intuitively demonstrate how the large selective kernel works by adaptively collecting the corresponding large receptive field for different objects.

\section{Experiments}

\subsection{Datasets}
HRSC2016~\cite{HRSC2016} is a high-resolution remote sensing images which is collected for ship detection. It consists of 1,061 images which contains 2,976 instances of ships. 

DOTA-v1.0~\cite{dota_set} consists of 2,806 remote sensing images. It contains 188,282 instances of 15 categories: Plane (PL), Baseball diamond (BD), Bridge (BR), Ground track field (GTF), Small vehicle (SV), Large vehicle (LV), Ship (SH), Tennis court (TC), Basketball court (BC), Storage tank (ST), Soccer-ball field (SBF), Roundabout (RA), Harbor (HA), Swimming pool (SP), and Helicopter (HC). 

FAIR1M-v1.0~\cite{sun_fair1m_2022} is a recently published remote sensing dataset that consists of 15,266 high-resolution images and more than 1 million instances. It contains 5 categories and 37 sub-categories objects. 

\subsection{Implementation Details}
In our experiment, we report the results of the detection model on HRSC2016, DOTA-v1.0 and FAIR1M-v1.0 datasets. To ensure fairness, we follow the same dataset processing approach as other mainstream methods~\cite{xie_oriented_2021,han_align_2020,han_redet_2021}. 
\implus{More details can be found in SM.}
During our experiments, the backbones are first pretrained on the ImageNet-1K~\cite{imagenet} dataset and then finetuned on the target remote sensing benchmarks.
In ablation studies, we adopt the 100-epoch backbone pretraining schedule for experimental efficiency (Tab.~\ref{tab:ablation_lk_decomp},~\ref{tab:max_avg_pool},~\ref{tab:ablation_key_comps},~\ref{tab:frames_comp},~\ref{tab:lk_sm_comp}). We adopt a 300-epoch backbone pretraining strategy to pursue higher accuracy in main results (Tab.~\ref{tab:hrsc},~\ref{tab:dota10},~\ref{tab:fair}), similarly to \cite{xie_oriented_2021, han_align_2020, yang_r3det_nodate, cheng_anchor-free_2022}. In main results (Tab.~\ref{tab:hrsc},~\ref{tab:dota10}), the ``Pre.'' column stands for the dataset on which the networks/backbones are pretrained (IN: Imagenet~\cite{imagenet} dataset; CO: Microsoft COCO~\cite{coco} dataset; MA: Million-AID~\cite{maid} dataset). Unless otherwise stated, LSKNet is defaulting to be built within the framework of Oriented RCNN~\cite{xie_oriented_2021} due to its compelling performance and efficiency.
All the models are trained on the training and validation sets and tested on the testing set. 
Following~\cite{xie_oriented_2021}, we train the models for 36 epochs on the HRSC2016 datasets and 12 epochs on the DOTA-v1.0 and FAIR1M-v1.0 datasets, with the AdamW~\cite{adamw} optimizer. The initial learning rate is set to 0.0004 for HRSC2016, and 0.0002 for the other two datasets, with a weight decay of 0.05. We use 8 RTX3090 GPUs with a batch size of 8 for model training, and use a single RTX3090 GPU for testing. All the FLOPs we report in this paper are calculated with a 1024$\times$1024 image input.

\subsection{Ablation Study}
In this section, we report ablation study results on the DOTA-v1.0 test set to investigate its effectiveness.

\begin{table}[t]
\centering
\renewcommand\arraystretch{1.2} 
\resizebox{0.90\columnwidth}{!}{
\begin{tabular}{lccll} 
($k$, $d$) sequence                                                                         & RF  & Num. & FPS  &mAP~(\%) \\ 
\Xhline{3\arrayrulewidth}
\begin{tabular}[c]{@{}l@{}}(29, 1)\end{tabular}                               & 29 & 1  & 18.6 & 80.66\\
\begin{tabular}[c]{@{}l@{}}(5, 1) $\longrightarrow$ (7, 4)\end{tabular}                & 29   & 2   & \textbf{20.5} & \textbf{80.91}\\
\begin{tabular}[c]{@{}l@{}}(3, 1) $\longrightarrow$ (5, 2) $\longrightarrow$ (7, 3)\end{tabular}  & 29   & 3   & 19.2 & 80.77\\
\end{tabular}}
\vspace{-8pt}
\caption{ \textbf{The effects of the number of decomposed large kernels} on the inference FPS and mAP, given theoretical receptive field being 29. We adopt LSKNet-T backbones pretrained on ImageNet for 100 epochs. Decomposing the large kernel into two depth-wise kernels achieves the  best performance of speed and accuracy.} 
\label{tab:ablation_lk_decomp}
\vspace{-8pt}
\end{table}

\textbf{Large Kernel Decomposition.}
Deciding on the number of kernels to decompose is a critical choice for the LSK module. 
We follow Eq.~\eqref{eqn:dr} to configure the decomposed kernels.
The results of the ablation study on the number of large kernel decompositions when the theoretical receptive field is fixed at 29 are shown in Tab.~\ref{tab:ablation_lk_decomp}. 
It suggests that decomposing the large kernel into two depth-wise large kernels results in a good trade-off between the speed and accuracy, achieving the best performance in terms of both FPS (frames per second) and mAP (mean average precision).

\implus{\textbf{Receptive Field Size and Selection Type.}}
Based on our evaluations presented in Tab.~\ref{tab:ablation_lk_decomp}, we find that the optimal solution for our proposed LSKNet is to decompose the large kernel into two depth-wise kernels in series. Furthermore, Tab.~\ref{tab:ablation_key_comps} shows that excessively small or large receptive fields can hinder the performance of the LSKNet, and a receptive field size of approximately 23 is determined to be the most effective. In addition, our experiments indicate that the proposed spatial selection approach is more effective than channel attention (similarly in SKNet~\cite{li2019selective}) for remote sensing object detection tasks.

\textbf{Pooling Layers in Spatial Selection.}
We conduct experiments to determine the optimal pooling layers for spatial selection in remote sensing object detection, as reported in Tab.~\ref{tab:max_avg_pool}. The results suggest that using both max and average pooling in the spatial selection component of our LSK module provides the best performance without sacrificing inference speed.

\begin{table}[t]
\centering
\renewcommand\arraystretch{1.2} 
\resizebox{0.80\columnwidth}{!}{
\begin{tabular}{cccc|ccc} 
($k_1$, $d_1$) & ($k_2$, $d_2$) & CS & SS & RF & FPS & mAP~(\%)  \\ 
\Xhline{3\arrayrulewidth}

(3, 1)   & (5, 2)   &  -  &  -  &  11  & 22.1 & 80.80 \\ 
(5, 1)   & (7, 3)   &  -  &  -  &  23  & 21.7  & 80.94 \\
(7, 1)   & (9, 4)   &  -  &  -  &  39  &  21.3 & 80.84\\ 
\hline

(5, 1)   & (7, 3)   & $\checkmark$  &  -  &  23  &  19.6 & 80.57  \\ 
(5, 1)   & (7, 3)   &  -  & $\checkmark$  &   23 & 20.7 & \textbf{81.31}   \\ 
\end{tabular}}
\vspace{-8pt}
\caption{\textbf{The effectiveness of the key design components} of the LSKNet when the large kernel is decomposed into a sequence of two depth-wise kernels. CS: channel selection (likewise in SKNet~\cite{li2019selective}); SS: spatial selection \textbf{(ours)}. We adopt LSKNet-T backbones pretrained on ImageNet for 100 epochs.  The LSKNet achieves best performance when using a reasonably large receptive field with spatial selection. 
}
\label{tab:ablation_key_comps}
\vspace{-8pt}
\end{table}

\begin{table}[t]
\centering
\renewcommand\arraystretch{1.2} 
\setlength{\tabcolsep}{4mm}{
\resizebox{0.60\columnwidth}{!}{
\begin{tabular}{cc|c|c} 
\multicolumn{2}{c|}{Pooling} & \multirow{2}{*}{FPS} & \multirow{2}{*}{mAP~(\%)}  \\ 
\cline{1-2}
Max. & Avg.                  &                      &                       \\ 
\Xhline{3\arrayrulewidth}
$\checkmark$    &                       &          20.7            &      81.23                 \\
     & $\checkmark$                     &           20.7       &        81.12               \\
$\checkmark$ & $\checkmark$          &             20.7       &            \textbf{81.31}           \\
\end{tabular}}}
\vspace{-8pt}
\caption{Ablation study on the effectiveness of the \textbf{maximum and average pooling in spatial selection} of our proposed LSK module. We adopt LSKNet-T backbones pretrained on ImageNet for 100 epochs. Best result is obtained when using both.}
\label{tab:max_avg_pool}
\vspace{-8pt}
\end{table}

\begin{table}[t]
\renewcommand\arraystretch{1.2} 
\centering
\resizebox{0.80\columnwidth}{!}{
\begin{tabular}{l|cc} 
Framework~\textbackslash~mAP~(\%)      & ResNet-18 & $\star$ LSKNet-T \\ 
\Xhline{3\arrayrulewidth}
Oriented RCNN~\cite{xie_oriented_2021}    &  79.27  &  81.31~{\small\textbf{(+2.04)}}  \\
RoI Transformer~\cite{ding_learning_2019} &  78.32  &   80.89~{\small\textbf{(+2.57)}}      \\
\begin{tabular}[c]{@{}l@{}}
S$^2$A-Net~\cite{han_align_2020} \\\end{tabular} &  76.82  &    80.15~{\small\textbf{(+3.33)}}  \\
R3Det~\cite{yang_r3det_nodate}       &   74.16   &  78.39~{\small\textbf{(+4.23)}}  \\ 
\hline
\#P {\scriptsize(backbone only)}  &    ~11.2M    &  ~~4.3M \small\textbf{{(-62\%)} }       \\
FLOPs {\scriptsize(backbone only)}  &    38.1G   &    19.1G \small\textbf{{(-50\%) }}       \\
\end{tabular}}
\vspace{-8pt}
\caption{\textbf{Comparison of LSKNet-T and ResNet-18} as backbones with different detection frameworks on DOTA-v1.0. The LSKNet-T backbone is pretrained on ImageNet for 100 epochs. The lightweight LSKNet-T achieves significant higher mAP in various frameworks than ResNet-18. }
\label{tab:frames_comp}
\vspace{-8pt}
\end{table}

\begin{table}[t]
\renewcommand\arraystretch{1.2} 
\centering
\resizebox{0.94\columnwidth}{!}{
\begin{tabular}{c|lccc} 
Group                                                & Model {\scriptsize(backbone only)}      & \#P   & FLOPs & mAP~(\%)  \\ 
\Xhline{3\arrayrulewidth}
Baseline & ResNet-18  &  11.2M   &   ~38.1G    &  79.27 \\
\hline
\multirow{3}{*}{\begin{tabular}[c]{@{}c@{}}Large \\ Kernel\end{tabular}}       
 & VAN-B1~\cite{guo_visual_2022}     & 13.4M  & ~52.7G  &  81.15    \\
& ConvNeXt V2-N~\cite{Woo2023ConvNeXtVC} &  15.0M   &   ~51.2G   & 80.81   \\
& MSCAN-S~\cite{guo_segnext_2022}   &  13.1M     &   ~45.0G    &   81.12   \\  
\hline
\multirow{3}{*}{\begin{tabular}[c]{@{}c@{}}Selective \\ Attention\end{tabular}} & SKNet-26~\cite{li2019selective}    & 14.5M & ~58.5G &  80.67\\  
 & ResNeSt-14~\cite{resnest}    &   ~8.6M  &   ~57.9G   &  79.51    \\     
  & SCNet-18~\cite{liu_improving_2020}      &   14.0M   &  ~50.7G     &    79.69  \\
  
\hline
\textbf{Ours}                                                        & $\star$ LSKNet-S      &    14.4M   &  ~54.4G    &  \textbf{81.48}   \\ 
\hline
\gc{Prev Best} & \gc{CSPNeXt}~\cite{lyu_rtmdet_2022} & \gc{26.1M} & ~\gc{87.6G} &  \gc{81.33} \\
\end{tabular}}
\vspace{-8pt}
\caption{\textbf{Comparison on LSKNet-S and other (large kernel/selective attention) backbones} under O-RCNN~\cite{xie_oriented_2021} framework on DOTA-v1.0, except that the \gc{Prev Best} is under RTMDet~\cite{lyu_rtmdet_2022} framework. All backbones are pretrained on ImageNet for 100 epochs. Our LSKNet achieves the best mAP under similar complexity budgets, whilst surpassing the previous best public records~\cite{lyu_rtmdet_2022}.
}
\label{tab:lk_sm_comp}
\vspace{-8pt}
\end{table}

\begin{table}[t]
\renewcommand\arraystretch{1.2} 
\centering
\setlength{\tabcolsep}{2mm}{
\resizebox{\columnwidth}{!}{
\begin{tabular}{lccccc} 
Method & Pre. & \textbf{mAP~(07) $\uparrow$} & \textbf{mAP~(12) $\uparrow$}  & \textbf{\#P $\downarrow$} & \textbf{FLOPs $\downarrow$}\\ 
\Xhline{3\arrayrulewidth}
DRN~\cite{pan_dynamic_2020}                  & IN       & -                         & 92.70       &  - &  -                \\
CenterMap~\cite{wang_learning_2021}          & IN       & -                       & 92.80      &  41.1M & 198G       \\
Rol Trans.~\cite{ding_learning_2019}        & IN    & 86.20                        & -        & 55.1M  & 200G    \\
G. V.~\cite{xu_gliding_2021}        & IN       & 88.20                        & -            & 41.1M  &  198G    \\
R3Det~\cite{yang_r3det_nodate}                 & IN       & 89.26                       & 96.01      &  41.9M  &  336G   \\
DAL~\cite{ming_dynamic_2021}                    & IN     & 89.77                       & -       &  36.4M & 216G    \\
GWD~\cite{yang_rethinking_2021}                    & IN      & 89.85                       & 97.37       &  47.4M & 456G  \\
S$^2$ANet~\cite{han_align_2020}              & IN     & 90.17                       & 95.01          & 38.6M  & 198G  \\
AOPG~\cite{cheng_anchor-free_2022}                   & IN      & 90.34                      & 96.22        &  - &  -  \\
ReDet~\cite{han_redet_2021} & IN   & 90.46 & 97.63   &  31.6M  & -     \\
O-RCNN~\cite{xie_oriented_2021}         & IN    & 90.50                        & 97.60               &  41.1M  & 199G  \\
RTMDet~\cite{lyu_rtmdet_2022}                 & CO     & 90.60                        & 97.10        & 52.3M  & 205G  \\
\hline
\rowcolor[rgb]{0.9,0.9,0.9}$\star$ LSKNet-S \textbf{(ours)}     & IN     &      \textbf{90.65}                            &        \textbf{98.46}         & \textbf{31.0M}  & \textbf{161G}   \\
\end{tabular}}}
\vspace{-8pt}
\caption{Comparison with state-of-the-art methods on the \textbf{HRSC2016} dataset. The LSKNet-S backbone is pretrained on ImageNet for 300 epochs, the same with most compared methods~\cite{yang_r3det_nodate,han_align_2020,xie_oriented_2021}. mAP (07/12): VOC 2007~\cite{voc2007}/2012~\cite{voc2012} metrics.}
\label{tab:hrsc}
\vspace{-8pt}
\end{table}

\begin{table*}[t]
\setlength{\tabcolsep}{4pt}
\renewcommand\arraystretch{1.2} 
\scriptsize      
\centering
\resizebox{\textwidth}{!}{
\begin{tabular}{l|c|c|c|c|ccccccccccccccc} 
Method                              &\textbf{Pre.}    & \textbf{mAP $\uparrow$} & \textbf{\#P $\downarrow$~~~} & \textbf{FLOPs $\downarrow$} & PL    & BD    & BR    & GTF   & SV    & LV    & SH    & TC    & BC    & ST    & SBF   & RA    & HA    & SP    & HC     \\ 
\Xhline{3\arrayrulewidth}
\textit{One-stage}                             &  &         &       &       &       &       &       &       &       &       &       &       &       &       &       &       &        \\ 
\hline
R3Det~\cite{yang_r3det_nodate}    & IN    & 76.47 & 41.9M & 336G & 89.80 & 83.77 & 48.11 & 66.77 & 78.76 & 83.27 & 87.84 & 90.82 & 85.38 & 85.51 & 65.57 & 62.68 & 67.53 & 78.56 & 72.62  \\
CFA~\cite{guo_beyond_2021}   & IN   & 76.67  & 36.6M & 194G & 89.08 & 83.20 & 54.37 & 66.87 & 81.23 & 80.96 & 87.17 & 90.21 & 84.32 & 86.09 & 52.34 & 69.94 & 75.52 & 80.76 & 67.96  \\
DAFNe~\cite{DAFNe}   & IN   & 76.95 &-&-  & 89.40 & \underline{86.27} & 53.70 & 60.51 & \underline{82.04} & 81.17 & 88.66 & 90.37 & 83.81 & 87.27 & 53.93 & 69.38 & 75.61 & 81.26 & 70.86  \\
SASM~\cite{SASM}   & IN   & 79.17 & 36.6M & 194G & 89.54 & 85.94 & 57.73 & 78.41 & 79.78 & 84.19 & 89.25 & 90.87 & 58.80 & 87.27 & 63.82 & 67.81 & 78.67 & 79.35 & 69.37  \\
AO2-DETR~\cite{dai_ao2-detr_2022} & IN   & 79.22 & 74.3M & 304G & 89.95 & 84.52 & 56.90 & 74.83 & 80.86 & 83.47 & 88.47 & 90.87 & 86.12 & \textbf{88.55} & 63.21 & 65.09 & 79.09 & \textbf{82.88} & 73.46  \\
S$^2$ANet~\cite{han_align_2020}     & IN  & 79.42  & 38.6M & 198G & 88.89 & 83.60 & 57.74 & 81.95 & 79.94 & 83.19 & \textbf{89.11} & 90.78 & 84.87 & 87.81 & 70.30 & 68.25 & 78.30 & 77.01 & 69.58  \\
R3Det-GWD~\cite{yang_rethinking_2021}   & IN    & 80.23 & 41.9M & 336G & 89.66 & 84.99 & 59.26 & 82.19 & 78.97 & 84.83 & 87.70 & 90.21 & 86.54 & 86.85 & \textbf{73.47} & 67.77 & 76.92 & 79.22 & 74.92  \\
RTMDet-R~\cite{lyu_rtmdet_2022}   & IN   & 80.54   & 52.3M & 205G & 88.36  & 84.96  & 57.33  & 80.46  & 80.58  & 84.88  & 88.08  & \textbf{90.90}  & 86.32  & 87.57     &  69.29     & 70.61  & 78.63  & 80.97  &  {79.24}\\
R3Det-KLD~\cite{yang_learning_2021}  & IN   & 80.63 & 41.9M & 336G & 89.92 & 85.13 & 59.19 & 81.33 & 78.82 & 84.38 & 87.50 & 89.80 & 87.33 & 87.00 & 72.57 & 71.35 & 77.12 & 79.34 & \underline{78.68}  \\
RTMDet-R~\cite{lyu_rtmdet_2022}   & CO    & 81.33  & 52.3M & 205G & 88.01 & 86.17 & 58.54 & 82.44 & 81.30 & 84.82 & 88.71 & {90.89} & \textbf{88.77} & 87.37 & 71.96 & 71.18 & 81.23 & 81.40 & 77.13  \\ 
\hline
\textit{Two-stage}                       &&      &   &         &       &       &       &       &       &       &       &       &       &       &       &       &       &       &        \\ 
\hline
SCRDet~\cite{yang_scrdet_2019}    & IN  & 72.61  & - & - & \underline{89.98} & 80.65 & 52.09 & 68.36 & 68.36 & 60.32 & 72.41 & 90.85 & 87.94 & 86.86 & 65.02 & 66.68 & 66.25 & 68.24 & 65.21  \\
Rol Trans.~\cite{ding_learning_2019}  & IN   & 74.61 & 55.1M &  200G & 88.65 & 82.60 & 52.53 & 70.87 & 77.93 & 76.67 & 86.87 & 90.71 & 83.83 & 82.51 & 53.95 & 67.61 & 74.67 & 68.75 & 61.03  \\
G.V.~\cite{xu_gliding_2021}    & IN  & 75.02   &  41.1M & 198G  & 89.64 & 85.00 & 52.26 & 77.34 & 73.01 & 73.14 & 86.82 & 90.74 & 79.02 & 86.81 & 59.55 & 70.91 & 72.94 & 70.86 & 57.32  \\
CenterMap~\cite{wang_learning_2021} & IN   & 76.03   & 41.1M & 198G  & 89.83 & 84.41 & 54.60 & 70.25 & 77.66 & 78.32 & 87.19 & 90.66 & 84.89 & 85.27 & 56.46 & 69.23 & 74.13 & 71.56 & 66.06  \\
CSL~\cite{yang_arbitrary-oriented_2020}   & IN    & 76.17   & 37.4M & 236G  & \textbf{90.25} & 85.53 & 54.64 & 75.31 & 70.44 & 73.51 & 77.62 & 90.84 & 86.15 & 86.69 & 69.60 & 68.04 & 73.83 & 71.10 & 68.93  \\
ReDet~\cite{han_redet_2021}   & IN   & 80.10  & 31.6M &  -  & 88.81 & 82.48 & 60.83 & 80.82 & 78.34 & 86.06 & 88.31 & 90.87 & \textbf{88.77} & 87.03 & 68.65 & 66.90 & 79.26 & 79.71 & 74.67  \\
DODet~\cite{DODet}   & IN  & 80.62   &  -&  - & 89.96 & 85.52 & 58.01 & 81.22 & 78.71 & 85.46 & 88.59 & {90.89} & 87.12 & 87.80 & 70.50 & 71.54 & 82.06 & 77.43 & 74.47  \\
AOPG~\cite{cheng_anchor-free_2022} & IN   & 80.66 &-& - & 89.88 & 85.57 & 60.90 & 81.51 & 78.70 & 85.29 & \underline{88.85} & {90.89} & 87.60 & 87.65 & 71.66 & 68.69 & 82.31 & 77.32 & 73.10  \\
O-RCNN~\cite{xie_oriented_2021}    & IN   & 80.87   & 41.1M &  199G & 89.84 & 85.43 & 61.09 & 79.82 & 79.71 & 85.35 & 88.82 & 90.88 & 86.68 & 87.73 & 72.21 & 70.80 & \underline{82.42} & 78.18 & 74.11  \\
KFloU~\cite{yang_kfiou_2022}  & IN   & 80.93   & 58.8M & 206G  & 89.44 & 84.41 & \underline{62.22} & 82.51 & 80.10 & \underline{86.07} & 88.68 & \textbf{90.90} & 87.32 & \underline{88.38} & \underline{72.80}  & \underline{71.95} & 78.96 & 74.95 & 75.27  \\ 
RVSA~\cite{wang_advancing_2022}  & MA & 81.24   & 114.4M & 414G & 88.97 & 85.76 & 61.46 & 81.27 & 79.98 & 85.31 & 88.30  & 90.84 & 85.06 & 87.50 & 66.77 & \textbf{73.11} & \textbf{84.75} & 81.88 & 77.58  \\ 
\hline
\rowcolor[rgb]{0.9,0.9,0.9}$\star$ LSKNet-T (\textbf{ours}) & IN &    {81.37}    & \textbf{21.0M} &  \textbf{124G}   &   89.14    &  84.90     & 61.78      & \underline{83.50}      & 81.54      & 85.87      & 88.64      & {90.89}      & 88.02      & 87.31      & 71.55      & 70.74      & 78.66      & 79.81      & 78.16       \\
\rowcolor[rgb]{0.9,0.9,0.9}$\star$ LSKNet-S (\textbf{ours}) & IN &   \underline{81.64}   & \underline{31.0M}  & \underline{161G}  &  89.57   &  \textbf{86.34} &   \textbf{63.13} & \textbf{83.67}  & \textbf{82.20}  &  \textbf{86.10} & 88.66 & {90.89}  & 88.41  & 87.42 & 71.72 & 69.58 & 78.88  & \underline{81.77} & 76.52  \\ 
\rowcolor[rgb]{0.9,0.9,0.9}$\star$ LSKNet-S* (\textbf{ours}) & IN &   \textbf{81.85}   & \underline{31.0M}  & \underline{161G}  &  89.69   &  {85.70} &   {61.47} & {83.23}  & {81.37}  &  {86.05} & 88.64 & {90.88}  & 88.49  & 87.40 & 71.67 & 71.35 & 79.19  & \underline{81.77} & \textbf{80.86}  \\ 

\end{tabular}}
\vspace{-8pt}
\caption{Comparison with state-of-the-art methods on the \textbf{DOTA-v1.0} dataset with multi-scale training and testing. The LSKNet backbones are pretrained on ImageNet for 300 epochs, similarly to the compared methods~\cite{yang_r3det_nodate,han_align_2020,xie_oriented_2021}. *: With EMA finetune similarly to the compared methods~\cite{lyu_rtmdet_2022}.} 
\label{tab:dota10}
\vspace{-4pt}
\end{table*}

\begin{table*}[t]
\renewcommand\arraystretch{1.2} 
\scriptsize 
\centering
\resizebox{0.96\textwidth}{!}{
\begin{tabular}{c|c|c|c|c|c|c|c|c} 
Model  & G. V.*~\cite{xu_gliding_2021} & RetinaNet*~\cite{retinanet} & C-RCNN*~\cite{cascade_rcnn} & F-RCNN*~\cite{frcnn}  & RoI Trans.*~\cite{ding_learning_2019}  & O-RCNN~\cite{xie_oriented_2021}  & {\cellcolor[rgb]{0.9,0.9,0.9}}$\star$ LSKNet-T & {\cellcolor[rgb]{0.9,0.9,0.9}}$\star$ LSKNet-S \\ 
\hline
\textbf{mAP(\%)}     & 29.92    & 30.67    & 31.18  & 32.12  & 35.29  & 45.60   & {\cellcolor[rgb]{0.9,0.9,0.9}} \underline{46.93}    & {\cellcolor[rgb]{0.9,0.9,0.9}} \textbf{47.87}     \\
\end{tabular}}
\vspace{-8pt}
\caption{Comparison with state-of-the-art methods on the \textbf{FAIR1M-v1.0} dataset.  The LSKNet backbones are pretrained on ImageNet for 300 epochs, similarly to~\cite{yang_r3det_nodate,han_align_2020,xie_oriented_2021}. *: Results are referenced from FAIR1M paper~\cite{sun_fair1m_2022}.}
\label{tab:fair}
\vspace{-8pt}
\end{table*}

\begin{table}[t]
\centering
\renewcommand\arraystretch{1.2} 
\resizebox{0.68\columnwidth}{!}{
\begin{tabular}{l|l|l} 
\multicolumn{1}{l|}{Team Name}                             & Pre-stage     & Final-stage      \\ 
\Xhline{3\arrayrulewidth}
nust\_milab  & 81.16 & 74.16  \\ 
\rowcolor[rgb]{0.9,0.9,0.9} Secret;Weapon \textbf{(ours)} & 81.11 & 73.94  \\ 
JiaNeng & 79.07 & 72.90  \\ 
ema.ai.paas                                       & 78.65 & 72.75  \\ 
SanRenXing & 78.06 & 71.39  \\
\end{tabular}}
\vspace{-8pt}
\caption{2022 the Greater Bay Area International Algorithm Competition results. The dataset is based on \textbf{FAIR1M-v2.0}~\cite{sun_fair1m_2022}.}
\label{tab:competition}
\vspace{-8pt}
\end{table}

\textbf{Performance of LSKNet backbone under different detection framworks.}
To validate the generality and effectiveness of our proposed LSKNet backbone, we evaluate its performance under various remote sensing detection frameworks, including two-stage frameworks O-RCNN~\cite{xie_oriented_2021} and RoI Transformer~\cite{ding_learning_2019} as well as one-stage frameworks S$^2$A-Net~\cite{han_align_2020} and R3Det~\cite{yang_r3det_nodate}.
The results in Tab.~\ref{tab:frames_comp} show that our proposed LSKNet backbone significantly improves detection performance compared to ResNet-18, while using only 38\% of its parameters and with 50\% fewer FLOPs.

\textbf{Comparison with Other Large Kernel/Selective Attention Backbones.}
We also compare our LSKNet with 6 \implus{popular} high-performance backbone models with large kernel or selective attention. As shown in Tab.~\ref{tab:lk_sm_comp}, under similar model size and complexity budgets, our LSKNet outperforms all other models on DOTA-v1.0 dataset.

\subsection{Main Results}
\textbf{Results on HRSC2016.} 
We evaluated the performance of our LSKNet against 12 state-of-the-art methods on the HRSC2016 dataset. The results presented in Tab.~\ref{tab:hrsc} demonstrate that our LSKNet-S outperforms all other methods with an mAP of \textbf{90.65\%} and \textbf{98.46\%} under the PASCAL VOC 2007~\cite{voc2007} and VOC 2012~\cite{voc2012} metrics, respectively. 

\textbf{Results on DOTA-v1.0.} 
We compare our LSKNet with 20 state-of-the-art methods on the DOTA-v1.0 dataset, as reported in Tab.~\ref{tab:dota10}. Our LSKNet-T and LSKNet-S achieve state-of-the-art performance with mAP of \textbf{81.37\%} and \textbf{81.64\%} respectively. Notably, our high-performing LSKNet-S reaches an inference speed of \textbf{18.1} FPS on 1024x1024 images with a single RTX3090 GPU.

\textbf{Results on FAIR1M-v1.0.} 
We compare our LSKNet against 6 other models on the FAIR1M-v1.0 dataset, as shown in Tab.~\ref{tab:fair}. The results reveal that our LSKNet-T and LSKNet-S perform exceptionally well, achieving state-of-the-art mAP scores of \textbf{46.93\%} and \textbf{47.87\%} respectively, surpassing all other models by a significant margin.

\begin{figure*}[t]
\begin{center}
\includegraphics[width=0.96\linewidth]{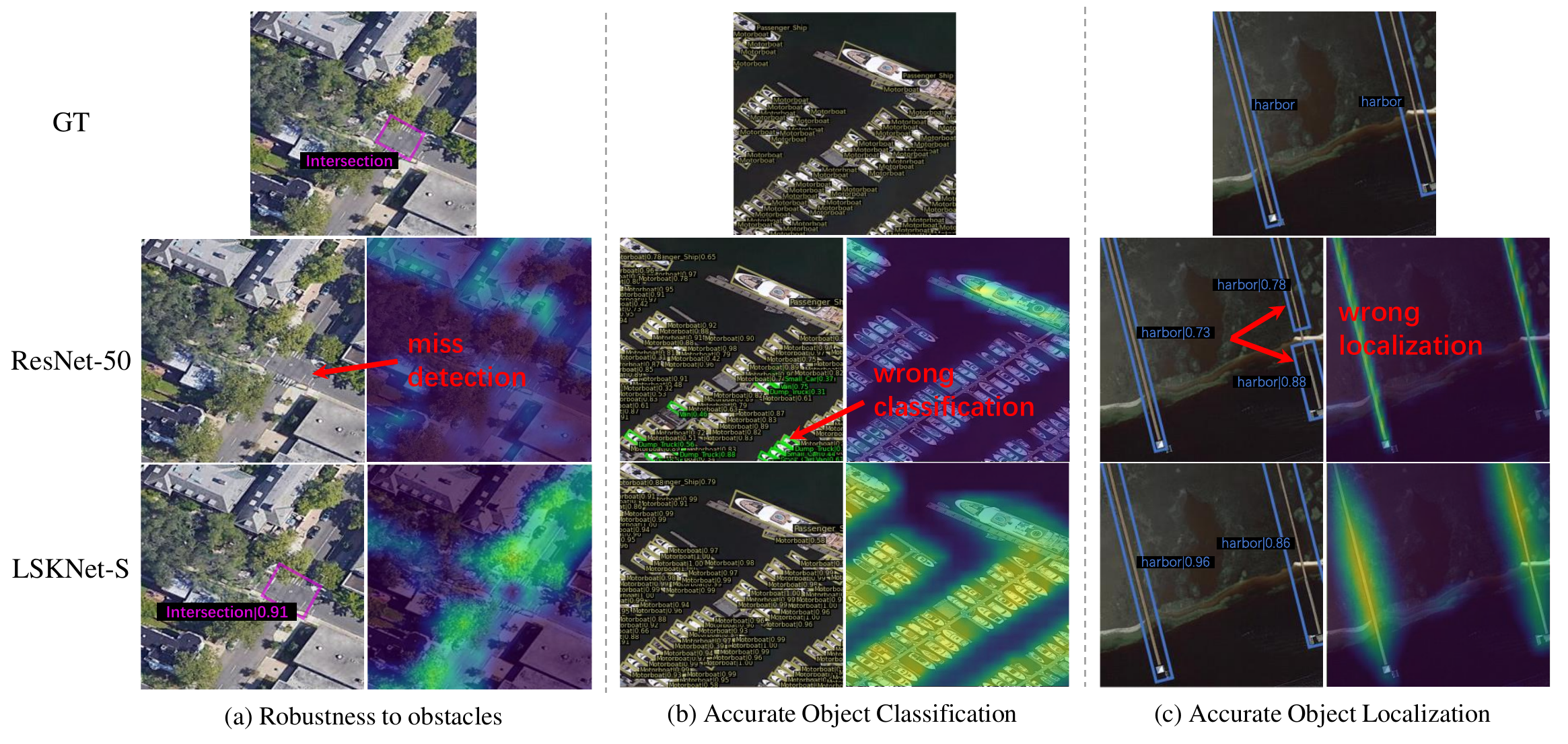}
\end{center}
    \vspace{-20pt}
   \caption{\textbf{Eigen-CAM visualization} of Oriented RCNN detection framework with ResNet-50 and LSKNet-S. Our proposed LSKNet can model a much long range of context information, leading to better performance in \implus{various} hard cases.}
   \vspace{-12pt}
\label{fig:cam}
\end{figure*}

\begin{figure}[t]
\begin{center}
\includegraphics[ width=0.82\columnwidth]{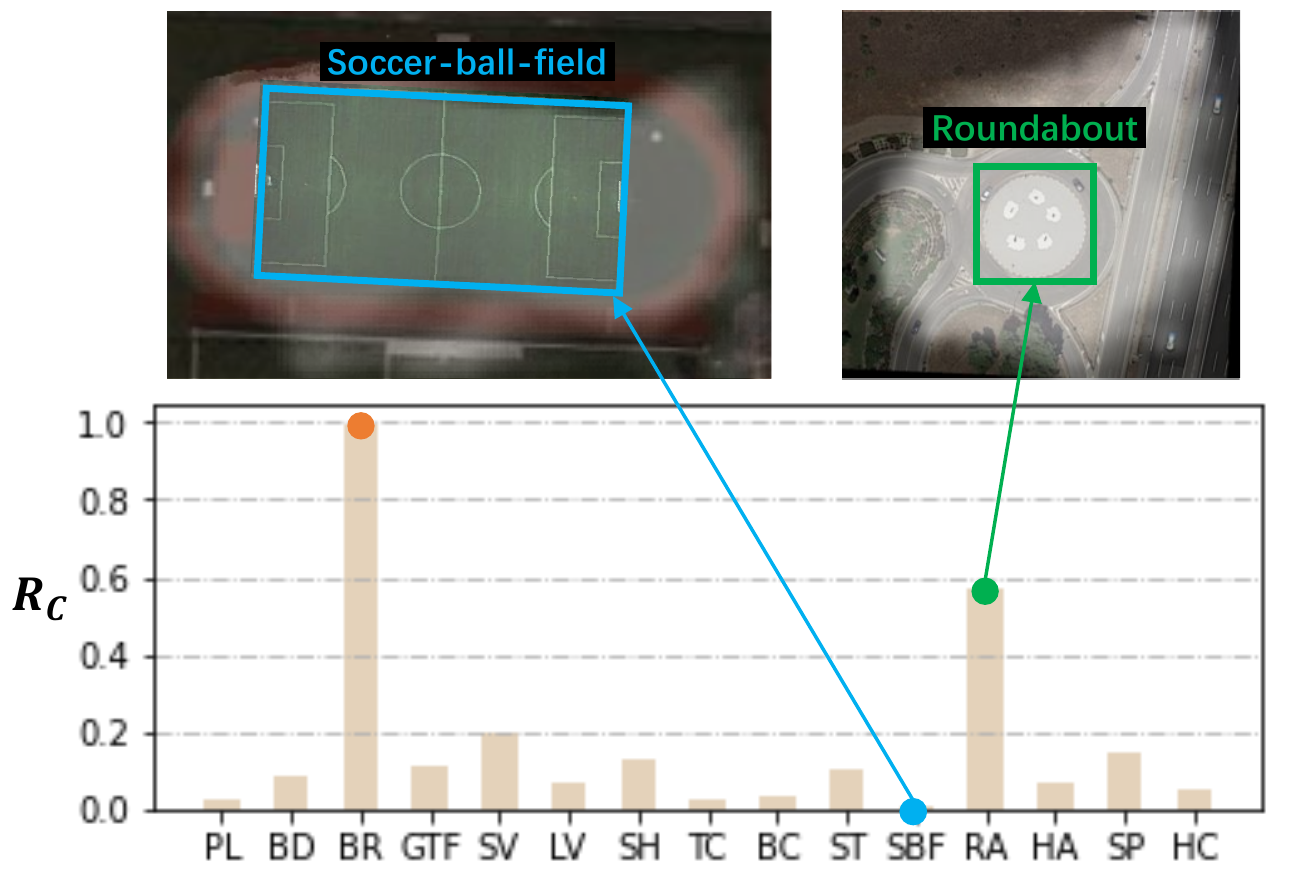}
\end{center}
    \vspace{-14pt}
   \caption{Normalised \textbf{Ratio ${R_c}$ of Expected Selective RF Area and GT Bounding Box Area} for object categories in DOTA-v1.0. The relative range of context required for different object categories varies a lot. Examples of Bridge and Soccer-ball-field are given, where the visualized receptive field is obtained from Eq.~\eqref{eqn:7} (i.e., the spatial activation) of our well-trained LSKNet model.}
\label{fig:ratio}
\vspace{-6pt}
\end{figure}

\textbf{2022 the Greater Bay Area International Algorithm Competition.}
Our team implemented a model similar to LSKNet for the \textit{2022 the Greater Bay Area International Algorithm Competition} and achieved second place, with a minor margin separating us from the first-place winner. The dataset used during the competition is a subset of FAIR1M-v2.0~\cite{sun_fair1m_2022}, and the competition results are illustrated in Tab.~\ref{tab:competition}. \implus{More details refer to SM.}

\subsection{Analysis}
Visualization examples of detection results and Eigen-CAM~\cite{eign_cam} are shown in Fig.~\ref{fig:cam}. It highlights that LSKNet-S can capture much more context information relevant to the detected targets, leading to better performance in various hard cases, which justifies our \textit{prior (1)}.

To investigate the range of receptive field for each object category, we define $R_c$ as the \textit{Ratio of Expected Selective RF Area and GT Bounding Box Area } for category $c$:
\vspace{-4pt}
\begin{equation}
   R_c = \frac{\sum_{i=1}^{I_c}{A_i / B_i}}{I_c} \text{,}
    \label{eqn:a_1}
\vspace{-10pt}
\end{equation}
\begin{equation}
    A_i = \sum_{d=1}^D{\sum_{n=1}^N{\lvert \widetilde{\mathbf{SA}}^d_n \cdot RF_n\rvert}},~ B_i = \sum_{j=1}^{J_i}{Area(\text{GT}_{j})} \text{,}
    \label{eqn:a_2}
\vspace{-4pt}
\end{equation}
\noindent where $I_c$ is the number of images that contain the object category $c$ only. The $A_i$ is the sum of spatial selection activation in all LSK blocks for input image $i$, where $D$ is the number of blocks in an LSKNet, and $N$ is the number of decomposed large kernels in an LSK module. $B_i$ is the total pixel area of all $J_i$ annotated oriented object bounding boxes (GT). 
\implus{We plot the normalized $R_c$ in Fig.~\ref{fig:ratio} which represents the relative range of context required for different object categories for a better view.}

\implus{The results suggest that the Bridge category stands out as requiring a greater amount of additional contextual information compared to other categories, primarily due to its similarity in features with roads and the necessity of contextual clues to ascertain whether it is enveloped by water. Conversely, the Court categories, such as Soccer-ball-field, necessitate minimal contextual information owing to their distinctive textural attributes, specifically the court boundary lines.}
It aligns with our knowledge and further supports \textit{prior (2)} that the relative range of contextual information required for different object categories varies greatly.

\begin{figure}[t]
\begin{center}
\includegraphics[width=0.87\columnwidth]{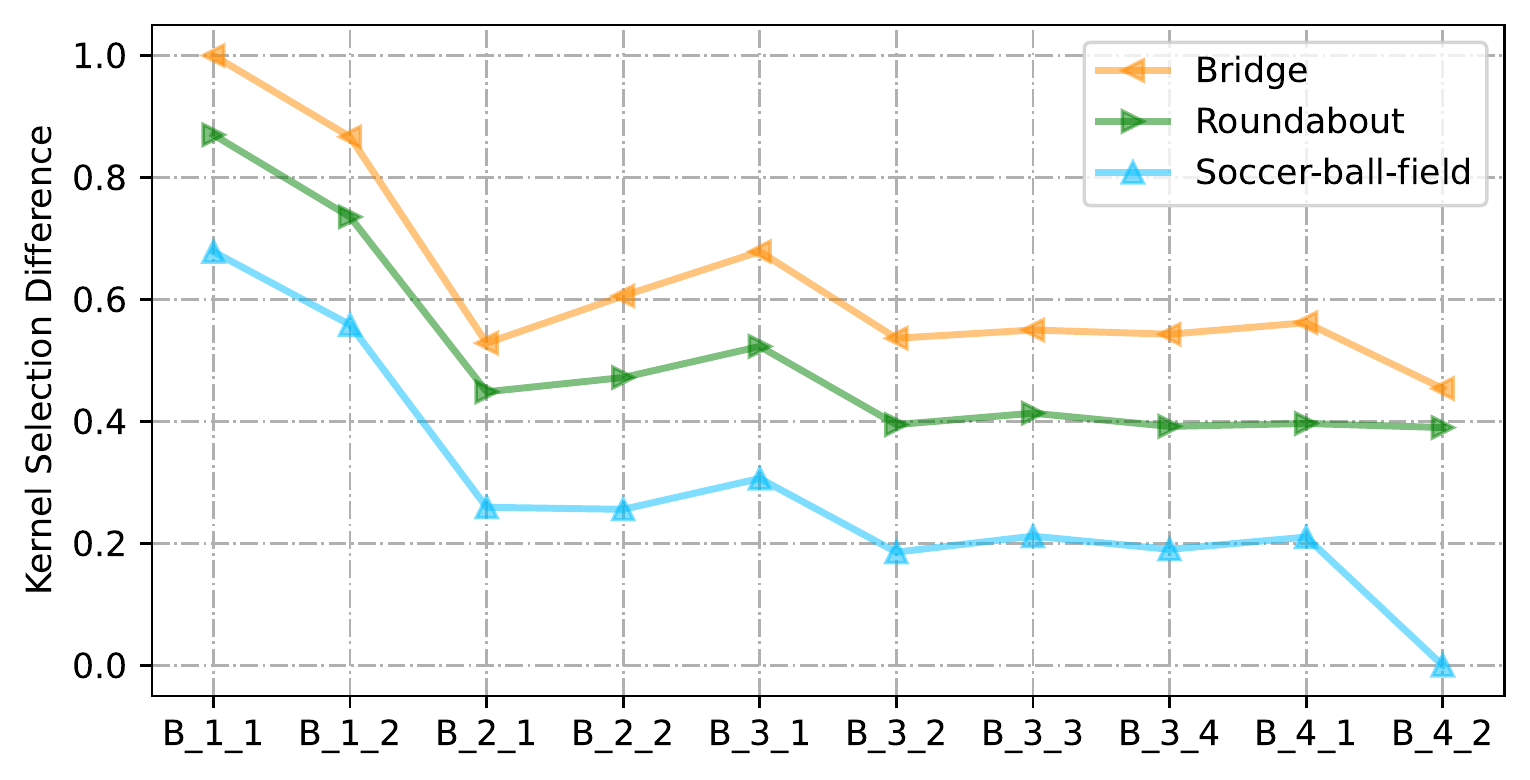}
\end{center}
    \vspace{-16pt}
   \caption{Normalised \textbf{Kernel Selection Difference} in the LSKNet-T blocks for Bridge, Roundabout and Soccer-ball-field. B\_i\_j represents the j-th LSK block in stage i. \implus{A greater value is indicative of a dependence on a broader context.}}
\label{fig:act_diff}
\vspace{-8pt}
\end{figure}

We further investigate the kernel selection behaviour in our LSKNet. For object category $c$, the \textit{Kernel Selection Difference} $\Delta A_c$ \implus{(i.e., larger kernel selection - smaller kernel selection)}
of an LSKNet-T block is defined as:
\vspace{-4pt}
\begin{equation}
    \Delta A_c = \lvert\widetilde{\mathbf{SA}}_{larger} - \widetilde{\mathbf{SA}}_{smaller}\rvert  \text{ .}
    \label{eqn:a_4}
    \vspace{-4pt}
\end{equation}
\implus{We demonstrate} the normalised $\Delta A_c$ over all images for \implus{three typical categories: Bridge, Roundabout and Soccer-ball-field and for each LSKNet-T block in Fig.~\ref{fig:act_diff}.} As expected, the participation of larger kernels of all blocks for Bridge is higher than that of Roundabout, and Roundabout is higher than Soccer-ball-field. This aligns with the common sense that Soccer-ball-field indeed does not require a large amount of context, since its own texture characteristics are already sufficiently distinct and discriminatory.

We also surprisingly discover another selection pattern of LSKNet across network depth: LSKNet usually utilizes larger kernels in its shallow layers and smaller kernels in higher levels. This indicates that networks tend to quickly focus on capturing information from large receptive fields in low-level layers so that higher-level semantics can contain sufficient receptive fields for better discrimination.


\section{Conclusion}
In this paper, we propose the Large Selective Kernel Network (LSKNet) for remote sensing object detection tasks, which is designed to utilize the inherent characteristics in remote sensing images: the need for a wider and adaptable contextual understanding. 
By adapting its large spatial receptive field, LSKNet can effectively model the varying contextual nuances of different object types. 
Extensive experiments demonstrate that our proposed lightweight model achieves state-of-the-art performance on the competitive remote sensing benchmarks.

{\small
\bibliographystyle{ieee_fullname}
\bibliography{egbib}
}

\clearpage

\appendix
\section{Appendix}

\subsection{LSKNet Block}
\begin{figure}[ht]
\begin{center}
\includegraphics[page=1, width=0.75\linewidth]{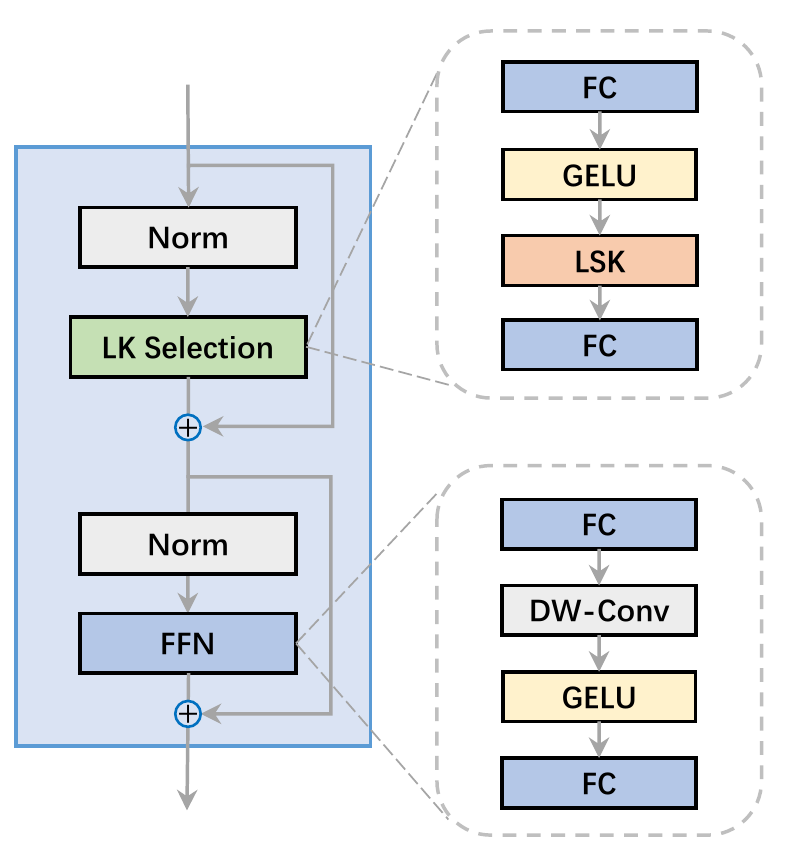}
\end{center}
    \vspace{-16pt}
   \caption{A block of LSKNet.}
\label{fig:lsk_blk}
\end{figure}

An illustration of an LSKNet Block is shown in Fig~\ref{fig:lsk_blk}. 
 The figure illustrates a repeated block in the backbone network, which is inspired by ConvNeXt~\cite{liu_convnet_2s022}, PVT-v2~\cite{PVT_v2}, VAN~\cite{guo_visual_2022}, Conv2Former~\cite{hou_conv2former_2022}, and MetaFormer~\cite{MetaFormer}. Each LSKNet block consists of two residual sub-blocks: the Large Kernel Selection (LK Selection) sub-block and the Feed-forward Network (FFN) sub-block. The LK Selection sub-block dynamically adjusts the network's receptive field as needed. The FFN sub-block is used for channel mixing and feature refinement which consists of a sequence of a fully connected layer, a depth-wise convolution, a GELU~\cite{gelu} activation, and a second fully connected layer. 

\subsection{2022 the Greater Bay Area International Algorithm Competition}
\begin{figure*}[ht]
\begin{center}
\includegraphics[page=1, width=0.85\linewidth]{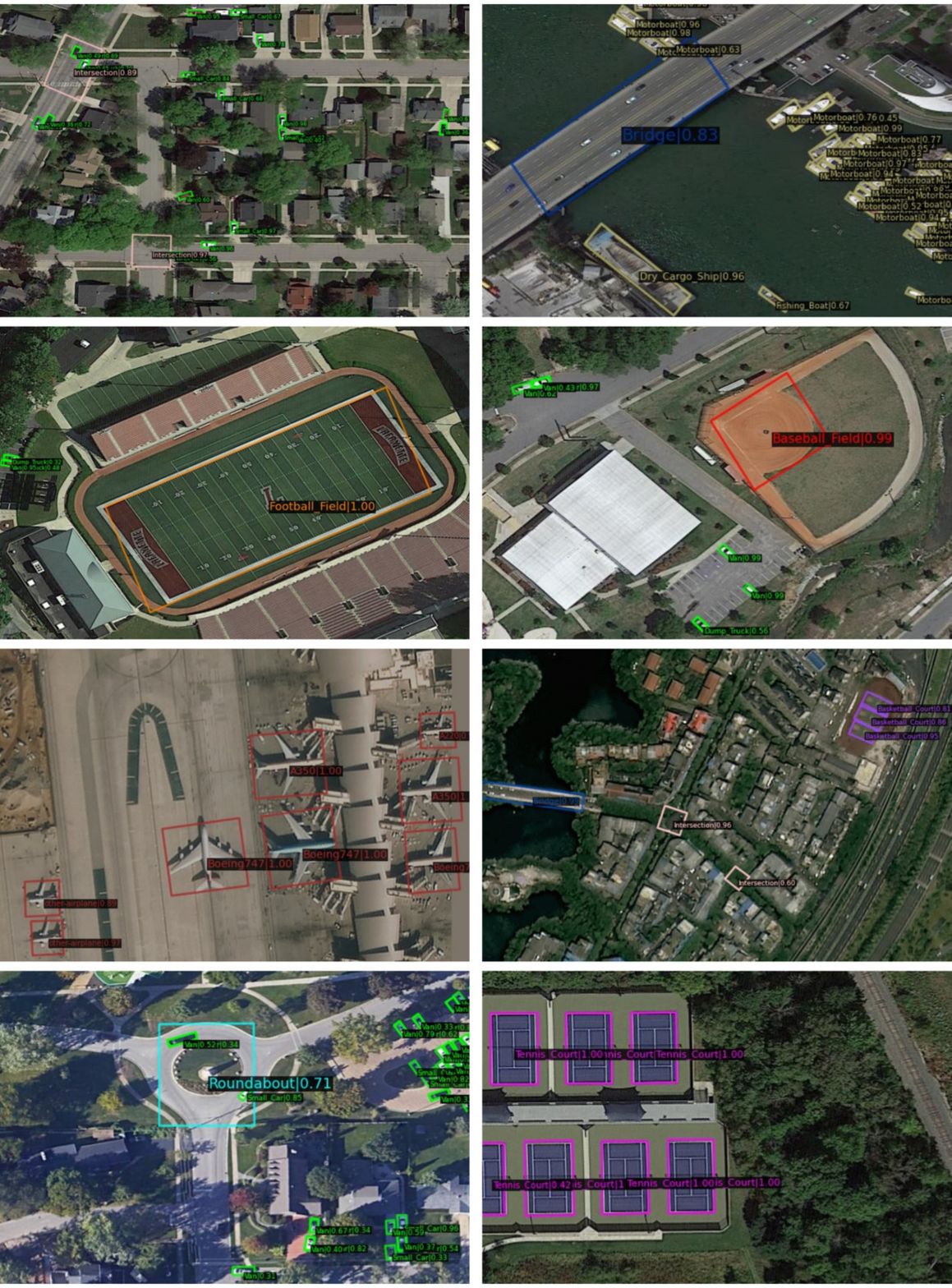}
\end{center}
    \vspace{-16pt}
   \caption{Examples of FAIR1M-v2.0 dataset test results with our LSKNet.}
\label{fig:fair1m15}
    
\end{figure*}

The competition requires participants to train a remote sensing image object detection model using the Jittor framework and produce rotated bounding boxes of objects and their respective types in test images. The dataset used for the competition is a subset of FAIR1M-v2.0 and is provided by the Chinese Academy of Sciences' Institute of Air and Space Information Innovation. It comprises 5000 training images, 576 preliminary test images, and 577 final test images. Example images of FAIR1M-v2.0 are shown in Fig.~\ref{fig:fair1m15}. The competition evaluates object detection performance based on ten object types: Airplane, Ship, Vehicle, Basketball\_Court, Tennis\_Court, Football\_field, Baseball\_field, Intersection, Roundabout, and Bridge. The mean Average Precision (mAP) evaluation metric is used, calculated based on the Pascal VOC 2012 Challenge. The Pre-stage and Final-stage are using the same finetuned model but with different test sets. The full competition scoreboard can be found at \url{https://www.cvmart.net/race/10345/rank}. 

In this competition, we employ model ensemble strategies to further enhance the performance of our single detection model. Two common methods for model ensemble in object detection are model output ensemble and model weight ensemble. Model output ensemble involves merging the outputs of different detectors using non-maximal suppression (NMS), while model weight ensemble merges the weights of multiple models into a single merged model through weighted averaging.
In order to achieve better results, we propose a multi-level ensemble strategy that combines both of these approaches. 
This strategy consists of two levels of ensembles. In the first level, we merge the weights of the two models with the best performance during training through weight averaging. In the second level, we merge the inference results of the two models using NMS. This forms a multi-layer ensemble mechanism that can produce the final ensembled inference results with high efficiency, using only two models.
By employing this multi-level ensemble strategy, we have achieved significant improvements in the performance of our object detection models in this competition as shown in Tab.~\ref{tab:model_ensemble}.
Some visualisation results of our proposed model on the FAIR1M-v2.0 test set are given in Fig.~\ref{fig:fair1m15}.

\begin{table}[t]
\centering
\renewcommand\arraystretch{1.2} 
\resizebox{0.7\columnwidth}{!}{
\begin{tabular}{lc} 
~~~~Model (Pre-stage)        & mAP~(\%) \\ 
\Xhline{3\arrayrulewidth}
Single model$_1$     & 80.29\\
Single model$_2$     & 80.42 \\ \hline
Output Ensemble                  & 80.51    \\
Weight Ensemble                  & 80.81    \\ \hline
\textbf{Multi-level Ensemble (ours)}             & 81.11   

\end{tabular}}
\caption{Multi-level model ensemble strategy results. } 

\label{tab:model_ensemble}
\vspace{-8pt}
\end{table}

\begin{figure*}[t]
    \begin{subfigure}[b]{\textwidth}
     \qquad
        \includegraphics[page=1, width=0.9\textwidth]{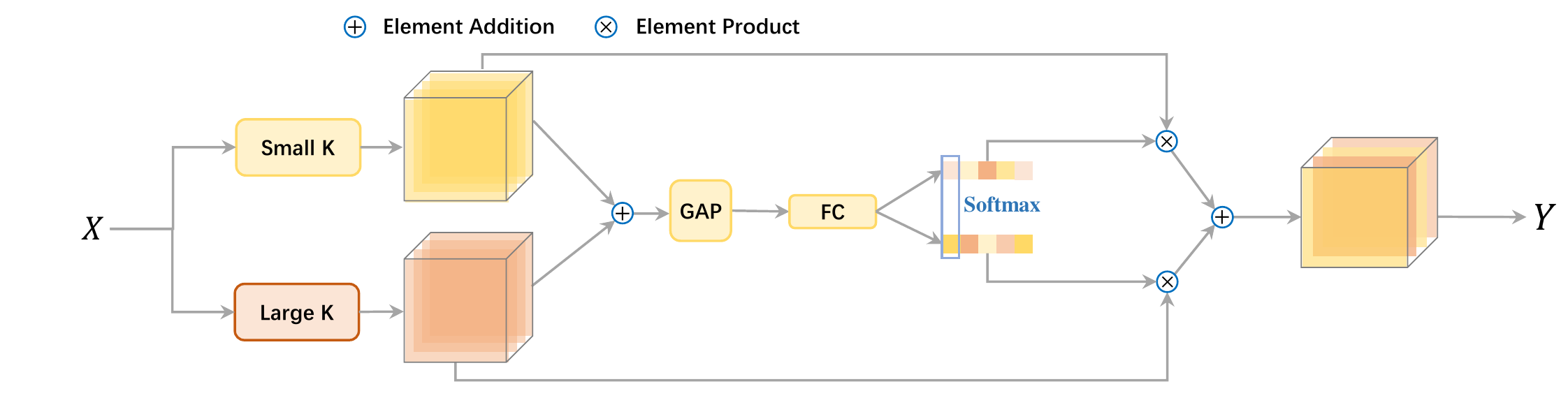}
        \caption{A conceptual illustration of SK module in \textbf{SKNet}.}
        \vspace{16pt}
    \end{subfigure}
  
    \begin{subfigure}[b]{\textwidth}
     \qquad
        \includegraphics[page=2, width=0.9\textwidth]{imgs/module_comp_cropped.pdf}
        \caption{A conceptual illustration of LSK module (with \textbf{Channel} Selection) in \textbf{LSKNet-CS}, which is corresponding to ``CS'' configuration in main paper Tab.~4.}
    \vspace{16pt}
    \end{subfigure}
    \begin{subfigure}[b]{\textwidth}
     \qquad
        \includegraphics[page=3, width=0.9\textwidth]{imgs/module_comp_cropped.pdf}
        \caption{A conceptual illustration of LSK module (with proposed \textbf{Spatial} Selection) in \textbf{LSKNet}.}
    \end{subfigure}
    
    \caption{Detailed conceptual comparisons between our proposed SKNet, LSKNet and LSKNet-CS.}
    \label{fig:sk_lsk_comp}
\end{figure*}

\begin{figure*}[t]
\begin{center}
\includegraphics[page=1, width=0.85\linewidth]{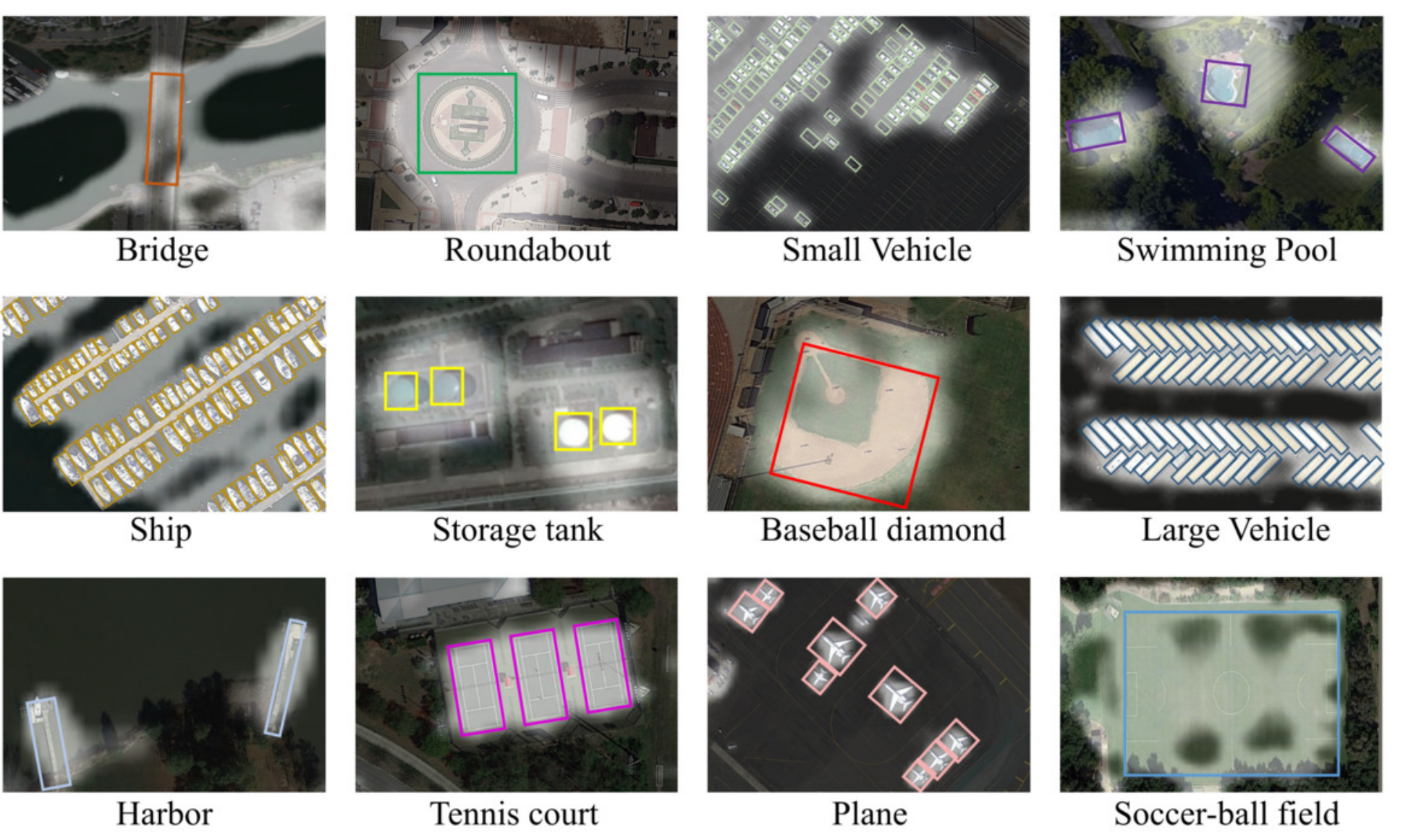}
\end{center}
    \vspace{-16pt}
   \caption{Receptive field activation for more object categories in DOTA-v1.0, where the activation map is obtained from the main paper Eq.~(8) (i.e., the spatial activation) of our well-trained LSKNet model. The object categories are arranged in decreasing order from top left to bottom right based on the \textit{Ratio of Expected Selective RF Area and GT Bounding Box Area } as illustrated in the main paper Fig.~6.}
\label{fig:more_act}
\end{figure*}

\subsection{SKNet v.s. LSKNet v.s. LSKNet-CS (channel selection version)}
A detailed conceptual comparison of SKNet, LSKNet and LSKNet-CS (channel selection version) module architecture is illustrated in Fig~\ref{fig:sk_lsk_comp}. {There are \textbf{two key distinctions} between SKNet and LSKNet. Firstly, our proposed selective mechanism relies explicitly on a sequence of large kernels via decomposition, a departure from most existing attention-based approaches. Secondly, our method adaptively aggregates information across large kernels in the spatial dimension, rather than the channel dimension as utilized by SKNet. This design is more intuitive and effective for remote sensing tasks, because channel-wise selection fails to model the spatial variance for different targets across the image space.}

\subsection{Experiment Implementation Details}
To ensure fairness, we follow the same dataset processing approach as other mainstream methods~\cite{xie_oriented_2021,han_align_2020,han_redet_2021}. For DOTA-v1.0 and FAIR1M-v1.0 datasets, we adopt multi-scale training and testing strategy by first rescaling the images into three scales (0.5, 1.0, 1.5), and then cropping each scaled image into 1024$\times$1024 sub-images with a patch overlap of 500 pixels. 
For the HRSC2016 dataset, we rescale the images by setting the longer side of the image to 800 pixels, without changing their aspect ratios.

\subsection{Spatial Activation Visualisations}

Receptive field activation examples for more object categories in DOTA-v1.0 are shown in Fig.~\ref{fig:more_act}, where the activation map is obtained from Eq.~(8) (i.e., the spatial activation) of our well-trained LSKNet model. It demonstrates that the Bridge category stands out as requiring a greater amount of additional contextual information compared to other categories, primarily due to its similarity in features with roads and the necessity of contextual clues to ascertain whether it is enveloped by water. Similarly, roundabouts also require a larger receptive field in order to distinguish between gardens and ring-like buildings. In order to accurately classify small objects such as ships and vehicles, a large receptive field is necessary to reference the surrounding context (i.e., whether it is in the sea or on land). Conversely, the Plane category and Court categories, such as Soccer-ball-field, necessitate minimal contextual information owing to their distinctive textural attributes, specifically the unique shapes and court boundary lines.


\subsection{FAIR1M benchmark results}
Fine-grained category result comparisons with state-of-the-art methods on the FAIR1M-v1.0 dataset are given in Tab.~\ref{tab:fair1m15}.

\begin{table*}[t]
\centering\renewcommand\arraystretch{1.2} 
\resizebox{0.95\textwidth}{!}{
\begin{tabular}{c|l|cccccccc}
Coarse Category & Sub Category        & \multicolumn{1}{l}{Gliding Vertex*}        & \multicolumn{1}{l}{RetinaNet*} & \multicolumn{1}{l}{Cascade RCNN*}          & \multicolumn{1}{l}{Faster RCNN*} & \multicolumn{1}{l}{ROI Trans*} & \multicolumn{1}{l}{Oriented RCNN} & \multicolumn{1}{l}{{\cellcolor[rgb]{0.9,0.9,0.9}}$\star$  LSKNet-T} & \multicolumn{1}{l}{{\cellcolor[rgb]{0.9,0.9,0.9}}$\star$  LSKNet-S}  \\ 
\Xhline{3\arrayrulewidth}
    \multirow{10}{*}{Airplane}            & Boeing737           & 35.43                                     & 38.46                         & 40.42                                     & 36.43                            & 39.58                         & 42.84                             & {\cellcolor[rgb]{0.9,0.9,0.9}}45.12                        & {\cellcolor[rgb]{0.9,0.9,0.9}}39.84                         \\
                & Boeing747           & 47.88                                     & 55.36                         & 52.86                                     & 50.68                            & 73.56                         & 87.61                             & {\cellcolor[rgb]{0.9,0.9,0.9}}84.97                        & {\cellcolor[rgb]{0.9,0.9,0.9}}86.63                         \\
                & Boeing777           & 15.67 & 24.75                         & 29.07                                     & 22.50                             & 18.32                         & 18.83                             & {\cellcolor[rgb]{0.9,0.9,0.9}}20.16                        & {\cellcolor[rgb]{0.9,0.9,0.9}}24.21                         \\
                & Boeing787           & 48.32                                     & 51.81                         & 52.47                                     & 51.86                            & 56.43                         & 62.92                             & {\cellcolor[rgb]{0.9,0.9,0.9}}56.00                        & {\cellcolor[rgb]{0.9,0.9,0.9}}56.48                         \\
                & C919                & ~~0.01                                      & ~~0.81                          & ~~0.00                                         & ~~0.01                             & ~~0.00                             & 22.17                             & {\cellcolor[rgb]{0.9,0.9,0.9}}25.77                        & {\cellcolor[rgb]{0.9,0.9,0.9}}24.17                         \\
        & A220                & 40.11                                     & 40.5                          & 44.37 & 47.81                            & 47.67                         & 47.87                             & {\cellcolor[rgb]{0.9,0.9,0.9}}50.05                        & {\cellcolor[rgb]{0.9,0.9,0.9}}52.20                         \\
                & A321                & 39.31                                     & 41.06                         & 38.35                                     & 43.83                            & 49.91                         & 70.25                             & {\cellcolor[rgb]{0.9,0.9,0.9}}71.63                        & {\cellcolor[rgb]{0.9,0.9,0.9}}73.31                         \\
                & A330                & 16.54                                     & 18.02                         & 26.55                                     & 17.66                            & 27.64                         & 73.34                             & {\cellcolor[rgb]{0.9,0.9,0.9}}67.94                        & {\cellcolor[rgb]{0.9,0.9,0.9}}72.82                         \\
                & A350                & 16.56                                     & 19.94                         & 17.54                                     & 19.95                            & 31.79                         & 77.19                             & {\cellcolor[rgb]{0.9,0.9,0.9}}74.04                        & {\cellcolor[rgb]{0.9,0.9,0.9}}75.83                         \\
                & ARJ21               & ~~0.01                                      & ~~1.70                           & ~~0.00                                         & ~~0.13                             & ~~0.00                             & 32.49                             & {\cellcolor[rgb]{0.9,0.9,0.9}}40.24                        & {\cellcolor[rgb]{0.9,0.9,0.9}}46.39                         \\ 
\hline
   \multirow{8}{*}{Ship}             & Passenger
  Ship    & ~~9.12                                      & ~~9.57                          & 12.10                                      & ~~9.81                             & 14.31                         & 20.21                             & {\cellcolor[rgb]{0.9,0.9,0.9}}19.23                        & {\cellcolor[rgb]{0.9,0.9,0.9}}20.43                         \\
                & Motorboat           & 23.34                                     & 22.55                         & 28.84                                     & 28.78                            & 28.07                         & 72.13                             & {\cellcolor[rgb]{0.9,0.9,0.9}}71.08                        & {\cellcolor[rgb]{0.9,0.9,0.9}}71.38                         \\
                & Fishing
  Boat      & ~~1.23                                      & ~~1.33                          & ~~0.71                                      & ~~1.77                             & ~~1.03                          & 13.53                             & {\cellcolor[rgb]{0.9,0.9,0.9}}14.70                        & {\cellcolor[rgb]{0.9,0.9,0.9}}15.81                         \\
                & Tugboat             & 15.67                                     & 16.37                         & 15.35                                     & 17.65                            & 14.32                         & 35.50                             & {\cellcolor[rgb]{0.9,0.9,0.9}}37.09                        & {\cellcolor[rgb]{0.9,0.9,0.9}}32.84                         \\
            & Engineering
  Ship  & 15.43                                     & 19.11                         & 18.53                                     & 16.47                            & 15.97                         & 16.23                             & {\cellcolor[rgb]{0.9,0.9,0.9}}16.60                        & {\cellcolor[rgb]{0.9,0.9,0.9}}14.79                         \\
                & Liquid
  Cargo Ship & 15.32                                     & 14.26                         & 14.63                                     & 16.19                            & 18.04                         & 26.49                             & {\cellcolor[rgb]{0.9,0.9,0.9}}24.74                        & {\cellcolor[rgb]{0.9,0.9,0.9}}25.37                         \\
                & Dry
  Cargo Ship    & 25.43                                     & 24.70                          & 25.15                                     & 27.06                            & 26.02                         & 38.43                             & {\cellcolor[rgb]{0.9,0.9,0.9}}40.57                        & {\cellcolor[rgb]{0.9,0.9,0.9}}41.29                         \\
                & Warship             & 13.56                                     & 15.37                         & 14.53                                     & 13.16                            & 12.97                         & 34.74                             & {\cellcolor[rgb]{0.9,0.9,0.9}}38.70                        & {\cellcolor[rgb]{0.9,0.9,0.9}}36.20                         \\ 
\hline
   \multirow{9}{*}{Vehicle}               & Small
  Car         & 66.23                                     & 65.20                          & 68.19                                     & 68.42                            & 68.80                          & 74.25                             & {\cellcolor[rgb]{0.9,0.9,0.9}}75.73                        & {\cellcolor[rgb]{0.9,0.9,0.9}}76.34                         \\
                & Bus                 & 23.43                                     & 22.42                         & 28.25                                     & 28.37                            & 37.41                         & 47.02                             & {\cellcolor[rgb]{0.9,0.9,0.9}}46.27                        & {\cellcolor[rgb]{0.9,0.9,0.9}}55.54                         \\
                & Cargo
  Truck       & 46.78                                     & 44.17                         & 48.62                                     & 51.24                            & 53.96                         & 50.22                             & {\cellcolor[rgb]{0.9,0.9,0.9}}54.06                        & {\cellcolor[rgb]{0.9,0.9,0.9}}55.84                         \\
                & Dump
  Truck        & 36.56                                     & 35.37                         & 40.40                                      & 43.60                             & 45.68                         & 57.56                             & {\cellcolor[rgb]{0.9,0.9,0.9}}59.52                        & {\cellcolor[rgb]{0.9,0.9,0.9}}61.57                         \\         & Van                 & 53.78                                     & 52.44                         & 58.00                                        & 57.51                            & 58.39                         & 75.22                             & {\cellcolor[rgb]{0.9,0.9,0.9}}75.57                        & {\cellcolor[rgb]{0.9,0.9,0.9}}76.71                         \\
                & Trailer             & 14.32                                     & 19.17                         & 13.66                                     & 15.03                            & 16.22                         & 20.91                             & {\cellcolor[rgb]{0.9,0.9,0.9}}19.30                        & {\cellcolor[rgb]{0.9,0.9,0.9}}21.46                         \\
                & Tractor             & 16.39                                     & ~~1.28                          & ~~0.91                                      & ~~3.04                             & ~~5.13                          & ~~2.99                              & {\cellcolor[rgb]{0.9,0.9,0.9}}~~3.68                         & {\cellcolor[rgb]{0.9,0.9,0.9}}~~7.19                          \\
                & Excavator           & 16.92                                     & 17.03                         & 16.45                                     & 17.99                            & 22.17                         & 19.95                             & {\cellcolor[rgb]{0.9,0.9,0.9}}28.40                        & {\cellcolor[rgb]{0.9,0.9,0.9}}25.73                         \\
                & Truck
  Tractor     & 28.91                                     & 28.98                         & 30.27                                     & 29.36                            & 46.71                         & ~~1.77                              & {\cellcolor[rgb]{0.9,0.9,0.9}}~~5.66                         & {\cellcolor[rgb]{0.9,0.9,0.9}}~~4.74                          \\ 
\hline
   \multirow{4}{*}{Court}               & Basketball
  Court  & 48.41                                     & 50.58                         & 38.81                                     & 58.26                            & 54.84                         & 55.35                             & {\cellcolor[rgb]{0.9,0.9,0.9}}59.74                        & {\cellcolor[rgb]{0.9,0.9,0.9}}61.78                         \\           & Tennis
  Court      & 80.31                                     & 81.09                         & 80.29                                     & 82.67                            & 80.35                         & 82.96                             & {\cellcolor[rgb]{0.9,0.9,0.9}}87.07                        & {\cellcolor[rgb]{0.9,0.9,0.9}}81.06                         \\
                & Football
  Field    & 53.46                                     & 52.50                          & 48.21                                     & 54.50                             & 56.68                         & 64.62                             & {\cellcolor[rgb]{0.9,0.9,0.9}}69.67                        & {\cellcolor[rgb]{0.9,0.9,0.9}}70.39                         \\
                & Baseball
  Field    & 66.93                                     & 66.76                         & 67.90                                      & 71.71                            & 69.07                         & 90.36                             & {\cellcolor[rgb]{0.9,0.9,0.9}}90.03                        & {\cellcolor[rgb]{0.9,0.9,0.9}}89.94                         \\ 
\hline
                & Intersection        & 59.41                                     & 60.13                         & 55.67                                     & 59.86                            & 58.44                         & 60.82                             & {\cellcolor[rgb]{0.9,0.9,0.9}}60.58                        & {\cellcolor[rgb]{0.9,0.9,0.9}}62.90                         \\
Road            & Roundabout          & 16.25                                     & 17.41                         & 20.35                                     & 16.92                            & 18.58                         & 20.47                             & {\cellcolor[rgb]{0.9,0.9,0.9}}23.20                        & {\cellcolor[rgb]{0.9,0.9,0.9}}27.00                         \\
                & Bridge              & 10.39                                     & 12.58                         & 12.62                                     & 11.87                            & 31.81                         & 33.40                             & {\cellcolor[rgb]{0.9,0.9,0.9}}38.57                        & {\cellcolor[rgb]{0.9,0.9,0.9}}39.51                         \\ 
\hline
 \multicolumn{2}{c|}{mAP (\%)}            & 29.92                                 & 30.67                         & 31.18                                  & 32.12                       & 35.29                  & 45.60                       & {\cellcolor[rgb]{0.9,0.9,0.9}} \underline{46.93}                         & {\cellcolor[rgb]{0.9,0.9,0.9}}\textbf{47.87  }                      
\end{tabular}}
\caption{Comparisons of fine-grained category results with state-of-the-art methods on the FAIR1M-v1.0 dataset. The LSKNet backbones are pretrained on ImageNet for 300 epochs, similarly to the compared methods R3Det~\cite{yang_r3det_nodate}, S2ANet~\cite{han_align_2020} and Oriented RCNN~\cite{xie_oriented_2021}. *: Results are referenced from FAIR1M~\cite{sun_fair1m_2022} paper.}
\label{tab:fair1m15}
\end{table*}

\end{document}